\documentclass[11pt,table,dvipsnames,numbers]{article}

\pdfoutput=1
\usepackage[preprint]{neurips_2024}
\usepackage[utf8]{inputenc}
\usepackage[T1]{fontenc}
\usepackage{hyperref}
\usepackage{url}
\usepackage{booktabs}
\usepackage{amsfonts}
\usepackage{nicefrac}
\usepackage{microtype}
\usepackage{xcolor}
\usepackage{times}
\usepackage{latexsym}
\usepackage{CJKutf8}
\usepackage{graphicx}
\usepackage{bigstrut}
\usepackage{multirow}
\usepackage{amsmath}
\usepackage[normalem]{ulem}
\usepackage{multicol}
\usepackage{tipa}
\usepackage{bm}
\usepackage[ruled,linesnumbered,vlined]{algorithm2e}
\usepackage{paralist}
\usepackage{arydshln}
\usepackage{verbatim}
\usepackage{csquotes}
\usepackage{xspace}
\usepackage{mdwlist}
\usepackage{subfigure}
\usepackage{makecell}
\usepackage{array}
\usepackage{enumitem}

\usepackage{colortbl}
\usepackage{dsfont}
\usepackage{tabularx}
\usepackage{bbm}
\usepackage{tikz}
\usepackage{pgfplots}
\usepackage{wrapfig}

\usepackage{listings}
\usepackage{amsmath}%
\usepackage{MnSymbol}%
\usepackage{wasysym}%
\usepackage{algpseudocode}
\usepackage[nameinlink,capitalise]{cleveref}
\newcolumntype{C}[1]{>{\centering\arraybackslash}p{#1}}
\newcolumntype{L}[1]{>{\arraybackslash}p{#1}}

\title{
\textit{Put Your Money Where Your Mouth Is:}\\
Evaluating Strategic Planning and Execution of LLM Agents in an Auction Arena
}

\newcommand{\method}{\textsc{AucArena}\xspace}

\newcommand{\ie}{\textit{i.e.}\xspace}
\newcommand{\eg}{\textit{e.g.}\xspace}

\makeatletter
\def\adl@drawiv#1#2#3{%
        \hskip.5\tabcolsep
        \xleaders#3{#2.5\@tempdimb #1{1}#2.5\@tempdimb}%
                #2\z@ plus1fil minus1fil\relax
        \hskip.5\tabcolsep}
\newcommand{\cdashlinelr}[1]{%
  \noalign{\vskip\aboverulesep
           \global\let\@dashdrawstore\adl@draw
           \global\let\adl@draw\adl@drawiv}
  \cdashline{#1}
  \noalign{\global\let\adl@draw\@dashdrawstore
           \vskip\belowrulesep}}
\makeatother

\author{Jiangjie Chen\textsuperscript{\rm$\spadesuit$}\thanks{Work done during Jiangjie Chen's internship at Allen Institute for AI.},
Siyu Yuan\textsuperscript{\rm$\spadesuit$},
Rong Ye\textsuperscript{\rm$\spadesuit$},
Bodhisattwa Prasad Majumder\textsuperscript{\rm$\heartsuit$},
Kyle Richardson\textsuperscript{\rm$\heartsuit$}\\
\textsuperscript{\rm$\spadesuit$}Fudan University\quad
\textsuperscript{\rm$\heartsuit$}Allen Institute for AI\\
\texttt{jjchen19@fudan.edu.cn, kyler@allenai.org}
}

\begin{document}

\maketitle

\renewcommand{\thefootnote}{\fnsymbol{footnote}}
\renewcommand{\thefootnote}{\arabic{footnote}}

\begin{abstract}

Recent advancements in Large Language Models (LLMs) showcase advanced reasoning, yet NLP evaluations often depend on static benchmarks.
Evaluating this necessitates environments that test strategic reasoning in dynamic, competitive scenarios requiring long-term planning.
We introduce \method, a novel evaluation suite that simulates auctions, a setting chosen for being highly unpredictable and involving many skills related to resource and risk management, while also being easy to evaluate.
We conduct controlled experiments using state-of-the-art LLMs to power bidding agents to benchmark their planning and execution skills. 
Our research demonstrates that LLMs, such as GPT-4, possess key skills for auction participation, such as budget management and goal adherence, which improve with adaptive strategies. 
This highlights LLMs' potential to model complex social interactions in competitive contexts.
However, variability in LLM performance and occasional outperformance by simpler methods indicate opportunities for further advancements in LLM design and the value of our simulation environment for ongoing testing and refinement. 
\footnote{Project page: \url{https://auction-arena.github.io}.}

\end{abstract}

\section{Introduction}
\label{sec:intro}

Being autonomous requires an agent to be able to make decisions independently, do complex reasoning and planning, and manage risk and resources, among many others~\citep{steels1995robots,franklin1996agent}. 
Large Language Models (LLMs) have proven to be able to solve a wide range of tasks, with the boundaries of what's possible being pushed every day \citep{wei2022emergent,bubeck2023sparks}. 
Despite the increasing view of these models as autonomous agents (\ie, LLM agents,~\citealp{wang2023survey,sumers2023cognitive,xi2023rise}), a crucial question remains:  
\textit{Can LLM agents effectively do sequential decision-making in dynamic environments to achieve their strategic objectives?}

While the potential is evident~\citep{Nakajima2023,SignificantGravitas2023autogpt}, these capabilities have yet to be rigorously evaluated.
Traditional reasoning and planning benchmarks in NLP~\citep{geva-etal-2021-aristotle,sakaguchi-etal-2021-proscript-partially,yuan-etal-2023-distilling} mostly assess agents in static contexts.
Yet, real-world scenarios demand that autonomous agents not merely respond to input but also have the ability to create long-term goals and plans, and continuously revise their decisions.
To bridge this gap, one recent line of research focuses on immersing agents in simulation environments that mimic real-world scenarios~\citep{wang-etal-2022-scienceworld,Park2023GenerativeAgents,Liu2023AgentBench,xie2024travelplanner}, ones that often focus on a targeted set of skills.
However, designing such simulations can require significant engineering effort, and doing fine-grained evaluation in these environments can often be a challenge.

In this work, we go beyond simple instructions and goal-oriented tasks to encompass long-term sequential decision-making strategies.
We emphasize developing environments characterized by the following properties:
\begin{inparaenum}[\it 1)]
    \item being dynamic and inherently unpredictable, requiring agents to be adaptive;
    \item involving limited resources, making the assets for competition scarce and the rewards highly contested;
    \item Being quantifiable, facilitating easy evaluation.
\end{inparaenum}
Motivated by the dynamics of \textit{auctions}, which are widely studied in multi-agent systems and game theory~\citep{laffont1997game,tuyls2007evolutionary}, we introduce \method.
Auctions offer a fertile ground for assessing strategic planning, resource allocation, risk management, and competitive behaviors. 
In \method, agents act as bidders in simulated auctions, grounding their strategic capabilities into execution. 
This environment enables quantifying an agent's performance using numerical metrics, such as profit.

\begin{figure*}[t]
    \centering
    \includegraphics[width=\linewidth]{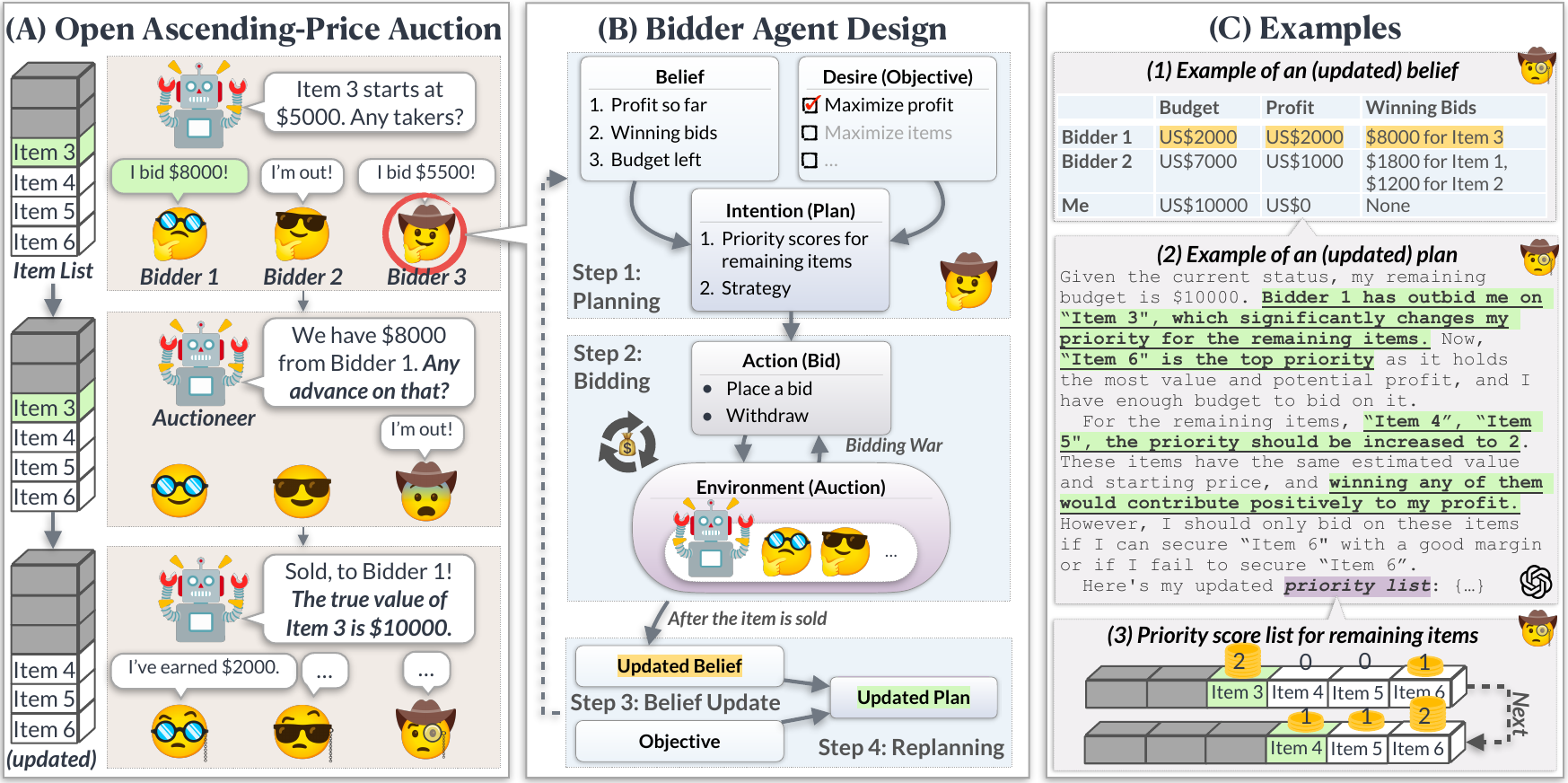}
    \caption{An illustration of \method: 
    (A) shows a multi-round, ascending-price auction with an auctioneer announcing the highest bids, where bidders publicly decide after private reasoning. 
    (B) presents a bidder agent structure using the Belief-Desire-Intention Model, involving planning, bidding, belief updating, and replanning, where beliefs and plans are adjusted with each auction development. 
    (C) offers instances of updated beliefs and plans, illustrating a bidder allocating priority scores to items post-reasoning.}
    \label{fig:front}
    \vspace{-0.3cm}
\end{figure*}

In \method, we set up an open auction where all bidders have equal information, as shown in Figure~\ref{fig:front}. 
Bidder agents engage in long-term sequential auctions with a fixed budget, allowing for strategic considerations.
Each agent monitors the environment and forms a plan to achieve its objective. 
To facilitate interaction with the auction environment, we incorporate four essential functions: planning, bidding, belief update, and replanning, each implemented with LLM prompting~\citep{wei2022chain,yao2022react}. 
These functions manifest the agents' strategic planning, adaptation to new information, and real-time decision-making capabilities.
We systematically explore the individual and collective behaviors of various LLM agents, examine their plan execution and adaptability, and delve into the dynamics of multi-objective multi-agent competitions.
Previous works and general language understanding tasks are more complex linguistically, while this work, though simpler in language, is more intricate due to the need for understanding and parsing object valuations and conducting game-theoretical analysis to make optimal decisions.

Our contributions include:
\begin{inparaenum}[\it 1)]
\item The creation of \method, an innovative simulated auction environment featuring a proposed bidder agent framework crafted to evaluate and measure the strategic decision-making skills of LLM agents within dynamic and competitive contexts.
\item The establishment of a new benchmark for evaluating the strategic performance of LLM agents, particularly emphasizing their ability to manage limited resources, engage in risk management, and adapt their strategies to achieve long-term objectives.
\item Utilizing \method, we gain insights into the strategic planning and execution skills of LLMs, emphasizing the importance of being adaptive in competitions. 
We show that even superior models like GPT-4 don’t always prevail in long-term strategic planning, and uncover new dynamics in multi-objective multi-agent competitions.
\end{inparaenum}

\section{Related Work}
\label{sec:related}
\textbf{Generative Agents}\quad
AI agents have evolved from simple, rule-based systems to interactive models and neural networks for dynamic responsiveness~\citep{mnih2013playing,silver2016mastering,silver2017mastering}. 
The recent development of \textit{generative LLM agents} has further broadened AI capabilities in natural language understanding and decision-making process.
\cite{Park2023GenerativeAgents} combined LLM with memory and reflection mechanisms, allowing for interaction between multiple language agents, and provided insights into how LLMs behave as agents in social simulation.
Additionally, building an LLM agent can also benefit from various prompting technique mechanisms~\citep{wei2022chain,yao2023tree}, which help in better decomposing complex tasks. Self-reflection or refinement mechanisms~\citep{yao2022react, shinn2023reflexion} enable the model to learn from feedback and improve.
Also, LLM agents can better accomplish complex tasks by utilizing external tools and calling APIs~\citep{schick2023toolformer,hao2023toolkengpt}.
In this work, we begin with the Belief-Desire-Intention model~\citep{bratman1987intention,georgeff1999belief} to construct an agent for a multi-item auction.

\textbf{Evaluation for Generative Agents}\quad
Recent works propose diverse interactive environments to evaluate the capabilities of LLM agents.
The existing environment include game-based environments~\citep{cote2019textworld, o2023hoodwinked, wang2023voyager,chalamalasetti2023clembench,duan2024gtbench}, web-based interactive environments~\citep{yao2022webshop, deng2023mindweb, zhou2023webarena, yang2023appagent}, and comprehensive measurement benchmarks~\citep{Liu2023AgentBench}.
In addition to evaluating the capabilities of a single agent, there are environments that provide a venue for multiple agents to interact and assess their abilities, such as cooperative tasks~\citep{chen2023agentverse,agashe2023evaluating,qian2023communicative} and competitive scenarios~\citep{zhao2023competeai,gu2024agent}.
Recent advancements have also focused on integrating strategic behaviors derived from game-theoretic principles~\cite{gemp2024states,jacob2024modeling,huang2024far}.
In this work, we leverage multi-agent auctions as a test-bed for assessing LLM agents' strategic planning, execution and adaptability in objective-driven environments with precise, measurable metrics.

\section{The Auction Arena}
\label{sec:method}

In this section, we introduce the implementation of our \method ($\mathsection$~\ref{sec:rule}) and agent architecture ($\mathsection$~\ref{sec:agent}) we use for our simulation.
We also introduce some details about our simulation design and implementation ($\mathsection$~\ref{sec:specification}).

\subsection{Open Ascending-Price Auction}
\label{sec:rule}
In this work, we adopt the English Auction where an auctioneer presents items in rounds and accepts ascending bids from participants until no higher bids are proposed. 
All bidders' actions are transparent and observable for fairness. 
The primary rule is that the highest bid wins the item.
Such a process is illustrated in Figure~\ref{fig:front}(A) and includes the following components: a list of items, an auctioneer, and bidders. An example auction log is provided in Appendix~\ref{appendix:log}.

\textbf{Items}\quad
Each item has a \textit{starting price} (\eg, \$1,000) for a bidding war, and a \textit{true value} (\$2,000) for resale.
We create a transparent bidding environment, omitting factors like personal or emotional value for simplicity.

\textbf{Auctioneer}\quad
The auctioneer, a rule-based agent, manages the bidding flow, declares the winners, and enforces auction rules, \eg, bidders cannot exceed their budgets, the next bid must surpass the prior highest bid by a minimum increase, etc.

\textbf{Bidders}\quad
Bidders are operationalized as LLM agents, each with their own strategies and assets.
They participate in the auction by bidding or withdrawing, influencing other bidders and thus shaping the dynamics of the auction and the outcome.
Formally, given an auction with $N$ bidders and $M$ items,
we use indicator $x_{i,j}=1$ to denote bidder~$i$ winning item~$j$, and $x_{i,j}=0$ otherwise.
The final bid price for item~$j$ is $p_j$, with its true value $v_j$. 
For simplicity, we assume that the true value of the item is identical for all bidders, but the bidders are \textbf{unaware} of the true value.
Bidders typically aim to maximize their profits in a multi-item auction but will be prompted to assume different objectives.
The utility function for a profit-driven bidder~$i$ is:
\begin{equation*}
    \textit{Maximize} \quad U_i = \sum_{j=1}^{M} (v_{j} - p_{j})\cdot x_{i,j}, 
    \quad \textit{s.t.}\ \  \sum_{j=1}^M p_j\cdot x_{i,j} \leq B_i, \ \ \sum_{i=1}^{N} x_{i,j}=1, x_{i,j}\in\{0,1\}
\end{equation*}
where $B_i$ is the budget for the bidder~$i$.
Note that the bidder will not incur any budget loss from a failed bid but may have negative profit if the winning bid exceeds its true value.
Also, bidders can pursue other objectives, \eg, securing a particular item or as many items as possible.

\subsection{Bidder Agent Architecture}
\label{sec:agent}

This section outlines the \textit{bidder agent architecture}, deploying the \textbf{Belief-Desire-Intention (BDI) Model}~\citep{bratman1987intention,georgeff1999belief,andreas-2022-language} to guide agent behavior and communication during auctions. The BDI model structures agents in terms of their:
\begin{inparaenum}[\it 1)]
    \item \textbf{Belief} - Knowledge of auction dynamics, such as budget, profits, and items won.
    \item \textbf{Desire} - Agent's auction objectives, primarily profit maximization or specific item acquisition.
    \item \textbf{Intention} - Agent's evolving strategy to achieve its desires, adaptable to new auction information.
\end{inparaenum}

This model underpins how agents strategize, adapt, and adjust, with auction outcomes influencing their beliefs and intentions for future auctions. 
We integrate four core actions within our agent architecture: \textit{planning}, \textit{bidding}, \textit{belief update}, and \textit{replanning}. 
Utilizing zero-shot prompts~\citep{kojima2022large}, we can interactively query LLM agents for insights into their strategies and beliefs, \eg, their understanding of their own budget, or the reasons why they made the decisions that they did, which enables a form of belief tracking~\citep{richardson2022breakpoint}. 
Details on instruction prompts are in Appendix~\ref{appendix:prompt}.

\textbf{Planning}\quad
Effective planning is crucial for agents to make informed decisions and well-thought-out bidding strategies that benefit both the present and future. 
This requires a bidding strategy in the initial step, where the agent~$i$ considers its budget~($B_i$) and all the available items, and generates a \textbf{pre-auction} textual plan. This plan acts as a strategic guide for efficient resource allocation throughout the multi-item auction.

\textbf{Bidding}\quad
In each round, non-leading bidders who are not the highest in the previous round can either place a higher bid or withdraw, while the previous top bidder skips bidding.
Initial bids start at or above the starting price, and by the final round, all but the winning bids are zero. 
In practice, we guide the agent to perform intermediate reasoning first before finalizing any decision.

\textbf{Belief Update}\quad
Context length restrictions prevent the inclusion of full bidding and conversation histories into LLMs. 
To compensate, we use a summarized dynamic memory of the auction's state, as depicted in Figure~\ref{fig:front}(C-1), which encompasses:
\begin{inparaenum}[\it 1)]
    \item \textbf{Remaining Budget},
    \item \textbf{Total Profit}, and
    \item \textbf{Winning Bids},
\end{inparaenum}
given in a JSON format. 
Belief updates for the agent occur post-bid to manage token limits. 
Errors in agent beliefs, such as misjudging items or profits, are recorded and corrected by the auctioneer (\textit{belief correction}) to facilitate uninterrupted gameplay, mirroring the use of notebooks and calculators by human bidders.

\textbf{Replanning}\quad
An important characteristic of \method is its dynamics and ever-changing nature, 
making earlier plans prone to becoming outdated and failing to execute.
Therefore, we add a replanning step for an agent to adjust its strategy based on the auction's progression and new information.
After bidding on an item, the agent reflects on its beliefs and previous plan and devises a new textual plan.
Then, the auctioneer moves the auction forward by presenting the next item, entering another bidding, belief update, and replanning iteration.

\subsection{Simulation Designs in \method}
\label{sec:specification}

Auctions can sometimes be difficult to understand due to numerous confounding factors.
Therefore, we propose three designs to simplify and facilitate further analysis while keeping the possibilities of more complex designs for future research.

\textbf{Starting Price and True Value of an Item}\quad
We set the actual value of an item in our simulation to be twice as its starting price, e.g., starting at \$1,000 while the true value is \$2,000.
Consequently, items with higher initial prices hold greater potential for profit if acquired, reflecting the principle that a higher investment and associated risk can lead to increased returns, which allows the study of intricate strategies of bidders.
The information of every artificial item is listed in Appendix~\ref{appendix:items}.

\textbf{Overestimation and Winner’s Curse}\quad
One intriguing aspect of auctions is the ``Winner’s Curse'', where the winning bid exceeds an item's true value, leading to losses — a common occurrence when bidders lack precise value estimations, a typical scenario in real auctions~\citep{kagel1986winner}.
To replicate this, we introduce an intentional overestimation of item value.
By default, we set the bidders to have an \textit{estimated value} that is 10\% higher than the \textit{true value}.
Bidders are not informed about items' true values but only estimate values.
This design allows us to study the risk management skills of LLM agents, where overbids can lead to apparent victories that are, in essence, strategic losses.

\textbf{Priority Score in the Plan}\quad
Consolidating the plans mentioned in $\mathsection$~\ref{sec:agent} is important to assess the capabilities of agents.
Simplifying, a bidder's future bidding strategy can be distilled into prioritizing remaining items.
When the auction moves to the $t$-th item (item $t$), bidder $i$ assigns a three-tier priority score for each remaining item $j$ ($t \leq j\leq M$), denoted as $r_{i,j}^{(t)} \in\{1,2,3\}$:
\begin{itemize}[itemsep=0pt, leftmargin=30pt, parsep=0pt, topsep=0pt, partopsep=-1pt]
    \item[\textbf{1 =}] The item is of minimal importance, and consider giving it up if necessary to save money;
    \item[\textbf{2 =}] The item holds value but isn't paramount, and could bid on it if budget allows;
    \item[\textbf{3 =}] The item is of utmost importance and is a top priority.
\end{itemize}
The scoring system, which is directly generated by LLMs, provides a tangible metric to gauge a bidder's foresight and adaptability.
As depicted in Figure~\ref{fig:front}(C-2,3), the agent first engages in intermediate reasoning of the current situation before making a structured plan.

\section{Experiments}
\label{sec:experiments}

\subsection{Experimental Setup}

We use the following state-of-the-art LLMs as the backend of bidding agents:
\begin{inparaenum}[\it 1)]
    \item \textbf{GPT-3.5-Turbo}~\citep{openai2022chatgpt}~(\texttt{gpt-3.5-turbo-1106});
    \item \textbf{GPT-4}~\citep{OpenAI2023GPT4TR}~(\texttt{gpt-4-0613}) and \textbf{GPT-4-Turbo}~(\texttt{gpt-4-1106-preview});
    \item \textbf{Gemini 1.0 Pro}~\citep{team2023gemini} by Google (\texttt{gemini-1.0-pro});
    \item \textbf{Mistral}~\citep{jiang2023mistral}, the 7 billion version (\texttt{mistral-7b-instruct-v0.2}) opensourced by MistralAI;
    \item \textbf{Mixtral}~\citep{mistralai_2023_mixtral} by MistralAI (\texttt{mixtral-8x7b-instruct-v0.1}), which is a Mixture-of-Expert (8 experts) version of Mistral (7b);
    \item \textbf{Rule Bidder}, as a baseline, has a fixed engagement limit per item, depending on the budget, and each of their bid increases the previous highest bid minimally (10\%), if a bid is possible.
\end{inparaenum}

We repeat each experiment by 10 runs for averaged performance, and set the temperature at $0$ for relatively deterministic behaviors.
Including input and output, the average length of each prompt is about 1500 to 1800 tokens, with the longest being around 5000 tokens.
The auctioneer requests a minimum bid increment at 10\% of the starting price to prevent protracted bidding, and bidders know the item order and starting prices throughout the auction.
In each auction game, by default, there are 2 items at each of the 5 starting prices, from \$1,000 to and \$5,000, totaling 10 items, so that the bidders will have to strategize for items of different values.

To objectively measure and rank the performance (\ie, total profit) of each bidder in such a competitive setting, we primarily employ the \textbf{TrueSkill score}~\citep{NIPS2006_f44ee263,minka2018trueskill}. 
The reason to use TrueSkill score is that it estimates dynamic skill levels ($\mu$) through Bayesian statistics while considering uncertainty ($\sigma$) in their true skills, which is commonly used in competitions such as online games or tournaments.

\begin{figure*}[t]
    \centering
    \includegraphics[width=\linewidth]{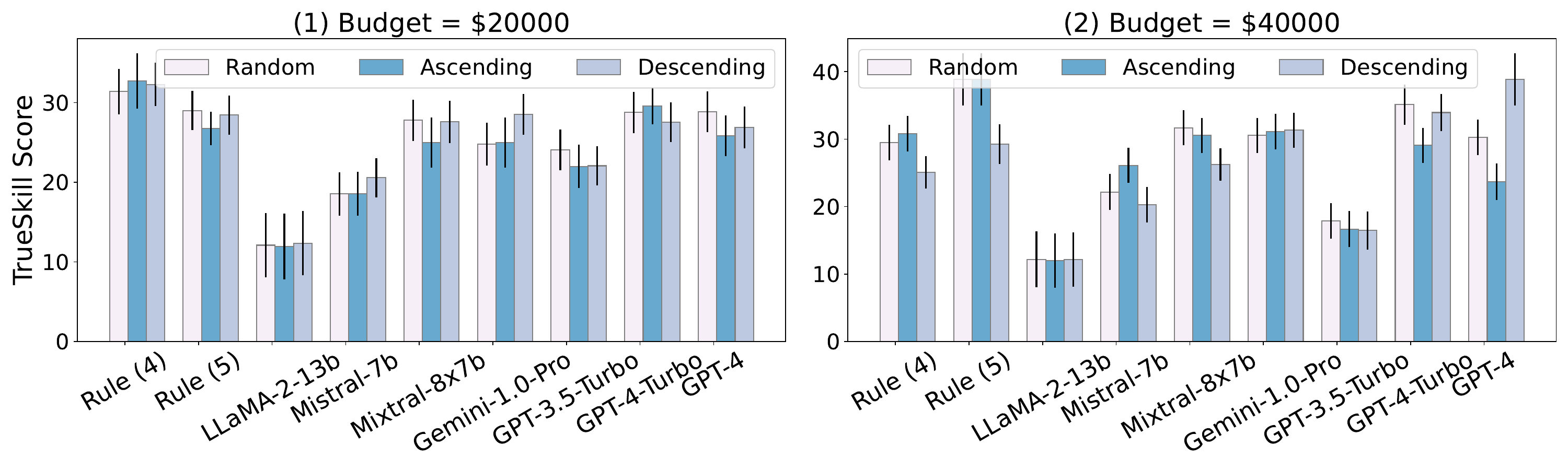}
    \caption{The TrueSkill scores of different bidder agents in the standard competition under 2 budget settings and 3 item orders based on items' starting prices.}
    \label{fig:main}
\end{figure*}

\subsection{Benchmarking LLMs as Bidders In Standard Competition}
\label{sec:benchmarking}

We set a three-bidder game as the \textit{Standard Competition} to benchmark bidder agents in \method, with 2 constant baseline participants and 1 challenger in the game.
By fixing two bidders and modifying the third, the evaluation is more standardized, ensuring that all LLMs are assessed within the same environment.
In this work, we use GPT-4-Turbo and GPT-3.5-Turbo as baselines, which are commonly studied by the community.
We design 6 (2$\times$3) settings in \method in total: 2 sets of budgets, \$20,000 and \$40,000;
and 3 types of item orders, including \textit{Random}, \textit{Ascending} and \textit{Descending} orders of starting prices.

 \begin{wrapfigure}{r}{0.45\linewidth}
    \centering
  \includegraphics[width=\linewidth]{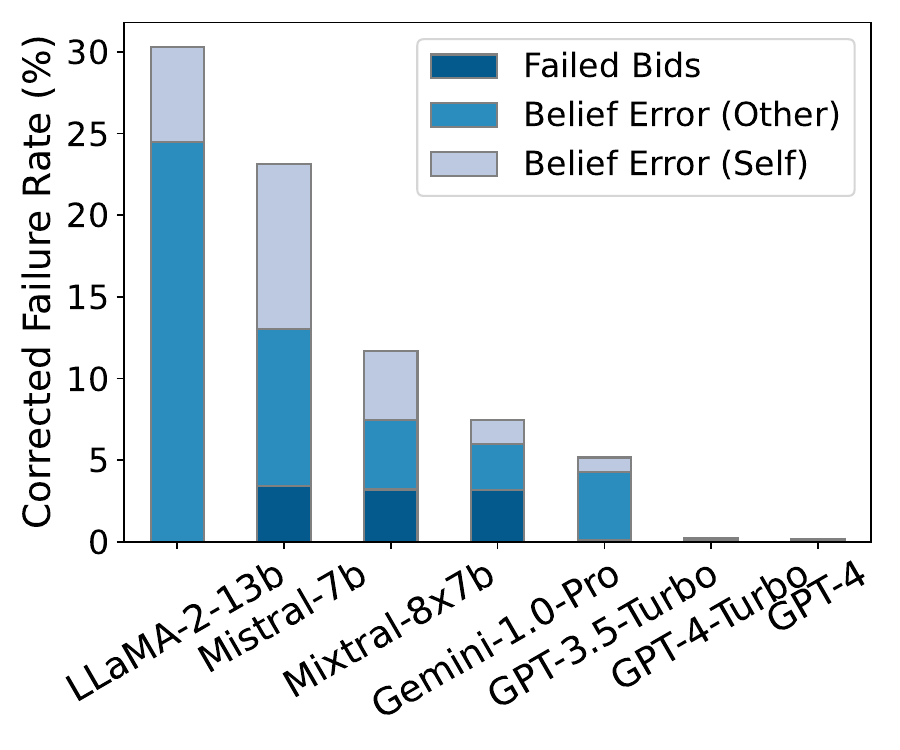}
  \caption{Different types of bidders' belief errors.}
  \label{fig:rationality}
 \end{wrapfigure}

According to Figure~\ref{fig:main}, a general trend of different agents can be observed with few variations.
GPT-4 exhibits higher scores than other LLM agents, suggesting it might be more effective in allocation efficiency or strategy under the given conditions; Gemini and Mixtral are also competitive in most settings.
Interestingly, their performance are not quite consistent when facing different item orders, \eg, GPT-4s are generally worse when facing items in a cheap-to-expensive order.
However, rule bidders establish an intriguing and powerful baseline, surpassing most LLMs in most settings.
In Appendix~\ref{appendix:human}, we demonstrate that, compared to GPT-4-Turbo, human bidders struggle when items are presented in a random order, but outperform others when the item order can be anticipated, whether it is descending or ascending.
We also showcase that GPT-3.5-Turbo frequently withdraws from the bidding war even if it has budget left for the items, whereas GPT-4-Turbo fails to employ sophisticated tactics to outmaneuver its competitors.

\textbf{Corrected Errors in Belief Tracking}\quad
Accurate and consistent monitoring of the auction environment is essential for success in dynamic auctions. 
Therefore, we track failed bids and incorrect beliefs of all participants as indicators of LLMs' state-tracking capabilities. 
Specifically:
\begin{inparaenum}[\itshape 1)]
    \item \textit{Failed Bids (\%)} measures the error rate in the bidding process, such as bids below the previous round's highest bid or beyond the budget;
    \item \textit{Belief Errors (\%)} assesses the accuracy in updating beliefs about the remaining budget, personal winning bids, and the status of other agents.
\end{inparaenum}
Notice that, the auctioneer corrects bidders' belief errors ($\mathsection$~\ref{sec:agent}) in \method, allowing us to concentrate on agents' planning and execution skills. 
We introduce the \textit{Corrected Failure Rate} ($\textit{CFR} = F/(C+F)$), where $F$ represents the number of failures and $C$ is the count of correct actions, a constant due to the eventual correction of failed actions.
As shown in Figure~\ref{fig:rationality}, GPT-4 demonstrates superior performance with the lowest error rates of 0.21\%, contrasting with other models that show higher factual error rates due to weaker context comprehension and arithmetic abilities. 
Note that, LLaMA-2-13b has no failed bids because it chooses to withdraw every time.

\subsection{Strategies and Behavioral Dynamics}
\label{sec:analysis}

To understand bidder agents' strategies and behaviors, we take a closer look at the bidding process of them, with a special focus on planning, execution, and faithfulness.

\begin{figure*}[t]
    \centering
    \small
    \includegraphics[width=\linewidth]{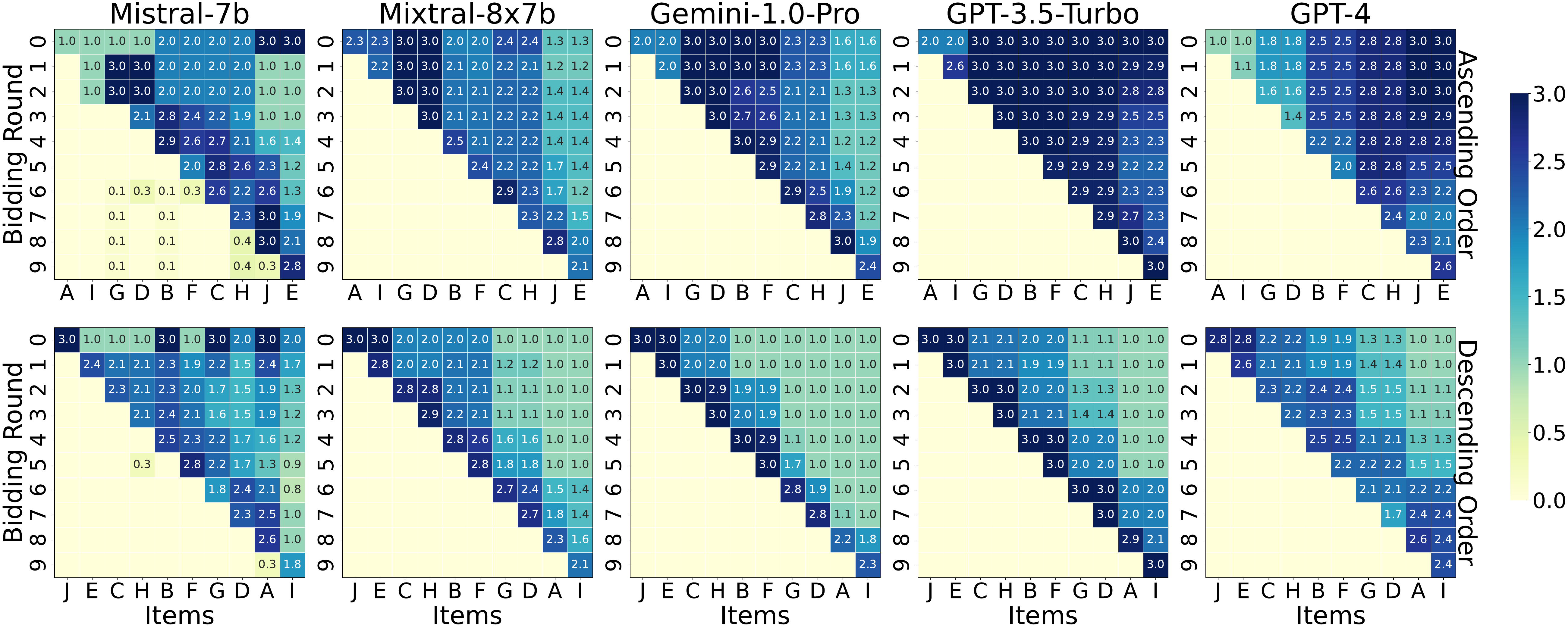}
    \caption{The heatmap of averaged priority scores and their changes before each bidding round under ascending (top) or descending (bottom) item orders, with a budget of \$20,000.
    }
    \label{fig:heatmap}
\end{figure*}

\textbf{Planning Analysis: Visualizing Long-term Strategies}\quad
Strategic behavior involves resource conservation for the future.
We present a visualization of item priority changes after each bidding round for bidder agents with a budget of \$20,000.
Figure~\ref{fig:heatmap} shows some interesting findings:
1) When items are arranged in ascending order (Figure~\ref{fig:bubble}(A)),
GPT-4 demonstrates restraint in the initial stages, opting to conserve budget for future opportunities, in contrast to other models that allocate their budget more quickly.
2) When items are arranged in descending order (Figure~\ref{fig:bubble}(B)), the majority of models exhibit a tendency to spend their budget more greedily, \ie, prioritize on expensive items. 
However, GPT-4 adopts a more cautious approach, refraining from early participation and instead reserves its budget for the subsequent 3-4 items.
3) There are some outliers in Mistral-7b, which are errors made by the agents, \ie, planning for the past items, not the future ones.
Appendix~\ref{appendix:heatmap} showcases a similar result when the budget is \$40,000.
Echoing the findings in Figure 4, in Appendix~\ref{appendix:bubble}, we also visualize the bidding behaviors of bidder agents that are concealed behind the final TrueSkill score in Standard Competition.
Evidence shows that GPT-4 tailors its strategy based on budget size and item order, showing increased competitiveness for higher-value items when the budget allows. 
Conversely, other models do not clearly demonstrate this adaptive behavior.

 \begin{wrapfigure}{r}{0.45\linewidth}
    \centering
    \includegraphics[width=\linewidth]{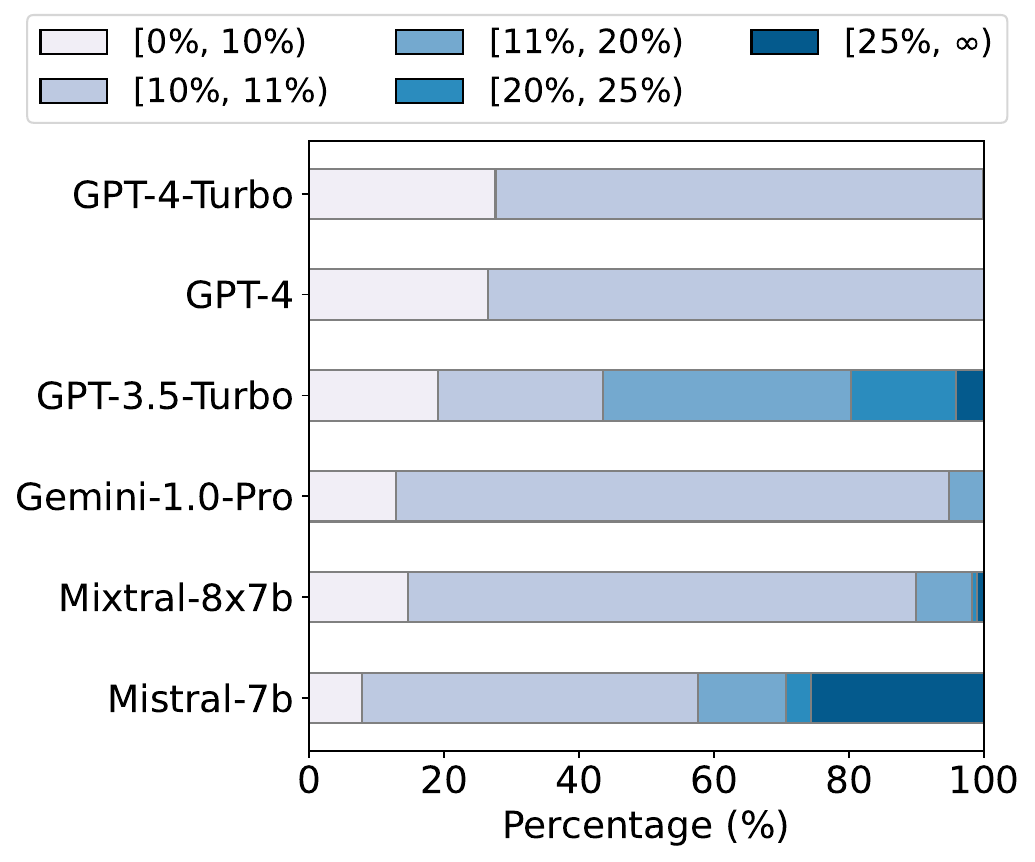}
    \caption{The Bid Increment Percentage for various agents, segmented into 5 different ranges. Each model's bar is divided to reflect the distribution of bids within these ranges.}
  \label{fig:aggressive}
 \end{wrapfigure}

\textbf{Execution Analysis: Bidding Aggressiveness}\quad
Assessing aggressiveness levels enables the anticipation and strategic planning of future bids to surpass competitors without considerable overpayment.
The Bid Increment Percentage (BIP) serves as a crucial metric for assessing competition-driven aggressiveness, indicating the percentile increase of a new bid over a bidder's preceding highest offer.
Ideally, in scenarios where the bid surpasses the previous by more than 10\%, cautious bidders strive to maintain their BIP as minimal as possible to evade the winner's curse. 
Figure~\ref{fig:aggressive} delineates the BIP distribution across various models. 
GPT-4s tend towards minimal bid increment, reflecting a conservative and rational bidding strategy to avoid excessive payments. 
Conversely, Mistral-7b and GPT-3.5-Turbo display a substantial volume of bids within the [11\%, $\infty$) increment bracket, suggesting a more aggressive, risk-tolerant bidding behavior. 
The range $[0, 10)$ denotes the increment percentage of the first bid on an item, which does not necessitate a 10\% increment from the starting price.

\begin{figure}[t]
    \centering
    \begin{minipage}{0.5\linewidth}
        \small
  \centering
    \label{tab:correlation}
    \begin{tabular}{lcccc}
    \toprule
        \multicolumn{1}{l}{\multirow{2}[0]{*}{\textbf{Model}}} & \multicolumn{2}{c}{\textbf{Initial ($r_{i,j}^{(1)}$)}} & \multicolumn{2}{c}{\textbf{Current ($r_{i,j}^{(j)}$)}} \\
        \cmidrule(lr){2-3}          
        \cmidrule(lr){4-5}
              & \multicolumn{1}{c}{\textbf{$\rho(r, n)$}} & \multicolumn{1}{c}{\textbf{$\rho(r, x)$}} & \multicolumn{1}{c}{\textbf{$\rho(r, n)$}} & \multicolumn{1}{c}{\textbf{$\rho(r, x)$}} \\
        \midrule
    LLaMA-2-13b & - & - & - & - \\
    Mistral-7b & .1252 & .2728 & .1139 & .2779 \\
    Mixtral-8x7b & .3553 & .4028 & .3994 & .4327 \\
    Gemini-1.0-Pro & .2457 & .1349 & .2776 & .2442 \\
    GPT-3.5-Turbo & .1849  & .0430  & .1494  & .0801  \\
    GPT-4 & .3741  & .3959  & .6318 & .6443  \\
    GPT-4-Turbo & .2348 & .0715 & .6423 & .4328 \\
    \bottomrule
\end{tabular}
\caption{Spearman correlation~$\rho(\cdot)$ between the priority score of an item in bidder~$i$'s initial or current plan, and the number of engagement~$n_{i,j}$ on item~$j$ and the winning indication~$x_{i,j}$. 
    LLaMA-2-13b is left blank due to its constant withdrawals.
    }
    \end{minipage}
    \hfill
 \begin{minipage}{0.47\linewidth}
        \includegraphics[width=\linewidth]{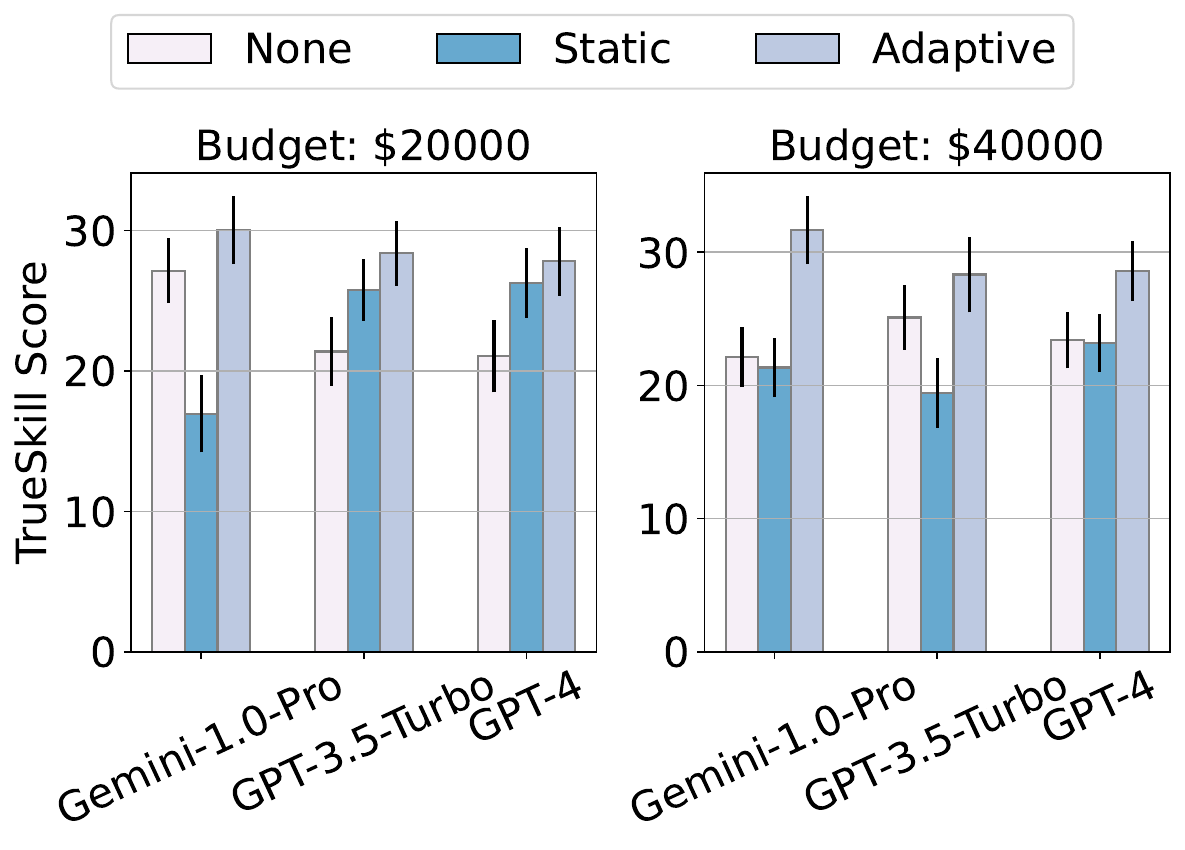} 
  \caption{The TrueSkill scores after competition between 3 agents powered by the same LLM, but with different modular designs.}
  \label{fig:modular_ablation}
    \end{minipage}
\vspace{-0.3cm}
\end{figure}

\begin{figure*}[t]
    \centering
    \small
    \includegraphics[width=\linewidth]{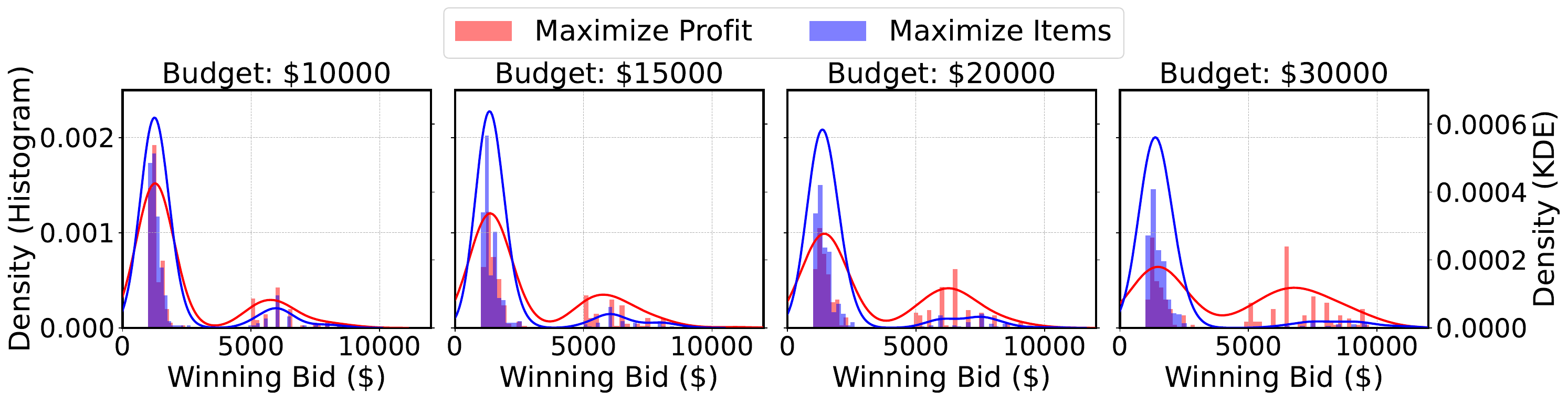}
    \caption{
    The competition between two groups of bidders: 2 profit bidders (red) and 2 item bidders (blue).
    We increase the budget of all bidders and show the changes in the histograms and kernel density estimation that depict the winning bids of two bidder groups.
    }
    \label{fig:niche}
\end{figure*}

\textbf{Do Agents Adhere to Their Plans during Execution?}\quad
To assess if LLMs follow and adapt their strategies during auctions, we analyze how their bidding behavior and assigned item priorities correlate, focusing on engagements and wins. 
We perform a correlation analysis between the priority scores bidders allocate to items and the auction outcomes. 
The priority score for an item is examined at two instances: at the beginning ($r_{i,j}^{(1)}$) and just before bidding on that item ($r_{i,j}^{(j)}$), so that we can check if the strategies evolve during auctions. 
Actions are quantified as the number of engagements (bids) on this item ($n_{i,j}$), and the indication of the winning bid $x_{i,j}\in\{0, 1\}$.
Spearman correlation coefficients, $\rho(r, n)$ and $\rho(r, x)$, help examine these relationships. 
Results in Table~\ref{tab:correlation} show that current plans correlate more with actions, highlighting the importance of adapting plans to the changing auction environment. 
There is noticeable variability in how models match their bids with their priorities over time. 
GPT-4 displays high correlations, indicating an effective adjustment in item priorities to deal with dynamic auction scenarios. 
Conversely, models like GPT-3.5-Turbo exhibit lower correlations, suggesting less flexible strategies or a weaker priority-action alignment.

\subsection{Modular Analysis}
\label{sec:ablation}
This experiment looks into the effects of architectural changes in LLMs and how they perform in environments with multiple competing objectives.

\textbf{Analysis on Planning and Replanning}\quad
In an ablation study focusing on the impact of different agent modules on LLM performance in auctions, we aim to pinpoint essential components for their success. 
We create a competitive setup with 3 agents, all powered by the same LLM, but each employing one of three distinct strategies:
\begin{inparaenum}[\itshape 1)]
    \item Adaptive Bidder: operates according to $\mathsection$~\ref{sec:agent};
    \item Static Bidder: lacks replanning module;
    \item None Bidder: lacks both planning and replanning modules.
\end{inparaenum}
According to Figure~\ref{fig:modular_ablation}, maintaining and updating plans enhances auction performance.
Relying on static plans sometimes yields worse outcomes than not planning at all, especially when the budget is \$40,000, as resources are abundant to facilitate greedy bidding (no plans).

\textbf{Analysis on Objectives: Emergent Niche Specification in Multi-Objective Competition}\quad
Just like ecosystems with diverse species in unique niches, multi-agent environments enable agents to pursue specialized objectives. 
In the context of auctions, participants often aim for goals beyond simply maximizing profit. 
We simulate two agent types:
\begin{inparaenum}[\itshape 1)]
    \item \textit{Profit Bidders}, focused on achieving high profit, and
    \item \textit{Item Bidders}, focused on acquiring the most items.
\end{inparaenum}
These objectives lead to different outcomes; for example, securing the most items doesn't equate to the highest profit, and vice versa. 
Profit Bidders are expected to target expensive, profitable items, whereas Item Bidders should seek out cheaper, more accessible items. 
To explore this, we divide 4 agents into 2 groups—2 Profit Bidders and 2 Item Bidders, including both GPT-4 and GPT-3.5-Turbo agents—and alter their budget across auctions for 20 items, including 4 expensive (starting at \$5,000) and 16 cheap items (starting at \$1,000). 
Figure~\ref{fig:niche} shows the density histograms and kernel density estimate plots of the winning bids, highlighting the niche specialization of the agents. 
As budgets increase to \$30,000, the differentiation between the two groups becomes more pronounced, with each specializing in either cheap or expensive items.
With a limited budget (\$10,000), Profit Bidders compete more for cheap items, giving Item Bidders opportunities to secure more profitable items.

\section{Conclusion}
\label{sec:conclusion}
In this paper, we propose \method to evaluate state-of-the-art LLMs as auction bidding agents, uncovering critical insights into the effectiveness of their strategies across various competitive auction scenarios. 
Through comprehensive analyses of strategies and behavioral dynamics, we demonstrate the importance of strategic planning and replanning capabilities and highlight the importance of tracking LLM agents' adaptive capabilities in changing environments. 
Further, ablation studies and the exploration of niche specialization offer valuable perspectives on the multifaceted roles LLMs can play in competitive scenarios. 
Our study advocates for further, innovative manipulations of our simulation method to explore these phenomena and amplify the potential of LLMs in modeling intricate social dynamics.

\section*{Acknowledgements}
We are grateful for the constructive comments from the anonymous reviewers. 
We thank the Aristo team at Allen Institute for AI for valuable feedback and support for this project, in particular: 
Peter Clark, Ashish Sabharwal, Tushar Khot, Nathaniel Weir, Oyvind Tafjord, Peter Hase and Sarah Wiegreffe.
We also thank Yao Fu from the University of Edinburgh, Hao Peng from UIUC, and Gregor Betz from the Karlsruhe Institute of Technology (KIT) for fruitful discussions that helped shape this project at various stages.
This work is supported by the Allen Institute for AI.

\bibliographystyle{unsrt}
\bibliography{neurips_2023}

\begin{thebibliography}{10}

\bibitem{steels1995robots}
Luc Steels.
\newblock When are robots intelligent autonomous agents?
\newblock {\em Robotics and Autonomous systems}, 15(1-2):3--9, 1995.

\bibitem{franklin1996agent}
Stan Franklin and Art Graesser.
\newblock Is it an agent, or just a program?: A taxonomy for autonomous agents.
\newblock In {\em International workshop on agent theories, architectures, and languages}, pages 21--35. Springer, 1996.

\bibitem{wei2022emergent}
Jason Wei, Yi~Tay, Rishi Bommasani, Colin Raffel, Barret Zoph, Sebastian Borgeaud, Dani Yogatama, Maarten Bosma, Denny Zhou, Donald Metzler, Ed~H. Chi, Tatsunori Hashimoto, Oriol Vinyals, Percy Liang, Jeff Dean, and William Fedus.
\newblock Emergent abilities of large language models.
\newblock {\em Transactions on Machine Learning Research}, 2022.
\newblock Survey Certification.

\bibitem{bubeck2023sparks}
S{\'e}bastien Bubeck, Varun Chandrasekaran, Ronen Eldan, Johannes Gehrke, Eric Horvitz, Ece Kamar, Peter Lee, Yin~Tat Lee, Yuanzhi Li, Scott Lundberg, et~al.
\newblock Sparks of artificial general intelligence: Early experiments with gpt-4.
\newblock {\em arXiv preprint arXiv:2303.12712}, 2023.

\bibitem{wang2023survey}
Lei Wang, Chen Ma, Xueyang Feng, Zeyu Zhang, Hao Yang, Jingsen Zhang, Zhiyuan Chen, Jiakai Tang, Xu~Chen, Yankai Lin, et~al.
\newblock A survey on large language model based autonomous agents.
\newblock {\em arXiv preprint arXiv:2308.11432}, 2023.

\bibitem{sumers2023cognitive}
Theodore Sumers, Shunyu Yao, Karthik Narasimhan, and Thomas~L Griffiths.
\newblock Cognitive architectures for language agents.
\newblock {\em arXiv preprint arXiv:2309.02427}, 2023.

\bibitem{xi2023rise}
Zhiheng Xi, Wenxiang Chen, Xin Guo, Wei He, Yiwen Ding, Boyang Hong, Ming Zhang, Junzhe Wang, Senjie Jin, Enyu Zhou, Rui Zheng, Xiaoran Fan, Xiao Wang, Limao Xiong, Yuhao Zhou, Weiran Wang, Changhao Jiang, Yicheng Zou, Xiangyang Liu, Zhangyue Yin, Shihan Dou, Rongxiang Weng, Wensen Cheng, Qi~Zhang, Wenjuan Qin, Yongyan Zheng, Xipeng Qiu, Xuanjing Huang, and Tao Gui.
\newblock The rise and potential of large language model based agents: A survey.
\newblock 2023.

\bibitem{Nakajima2023}
Yohei Nakajima.
\newblock Babyagi.
\newblock \url{https://github.com/yoheinakajima/babyagi}, 2023.

\bibitem{SignificantGravitas2023autogpt}
Significant-Gravitas.
\newblock Autogpt.
\newblock \url{https://github.com/Significant-Gravitas/Auto-GPT}, 2023.

\bibitem{geva-etal-2021-aristotle}
Mor Geva, Daniel Khashabi, Elad Segal, Tushar Khot, Dan Roth, and Jonathan Berant.
\newblock Did aristotle use a laptop? a question answering benchmark with implicit reasoning strategies.
\newblock {\em Transactions of the Association for Computational Linguistics}, 9:346--361, 2021.

\bibitem{sakaguchi-etal-2021-proscript-partially}
Keisuke Sakaguchi, Chandra Bhagavatula, Ronan Le~Bras, Niket Tandon, Peter Clark, and Yejin Choi.
\newblock pro{S}cript: Partially ordered scripts generation.
\newblock In {\em Findings of the Association for Computational Linguistics: EMNLP 2021}, pages 2138--2149, Punta Cana, Dominican Republic, 2021. Association for Computational Linguistics.

\bibitem{yuan-etal-2023-distilling}
Siyu Yuan, Jiangjie Chen, Ziquan Fu, Xuyang Ge, Soham Shah, Charles Jankowski, Yanghua Xiao, and Deqing Yang.
\newblock Distilling script knowledge from large language models for constrained language planning.
\newblock In {\em Proceedings of the 61st Annual Meeting of the Association for Computational Linguistics (Volume 1: Long Papers)}, pages 4303--4325, Toronto, Canada, July 2023. Association for Computational Linguistics.

\bibitem{wang-etal-2022-scienceworld}
Ruoyao Wang, Peter Jansen, Marc-Alexandre C{\^o}t{\'e}, and Prithviraj Ammanabrolu.
\newblock {S}cience{W}orld: Is your agent smarter than a 5th grader?
\newblock In {\em Proceedings of the 2022 Conference on Empirical Methods in Natural Language Processing}, pages 11279--11298, Abu Dhabi, United Arab Emirates, December 2022. Association for Computational Linguistics.

\bibitem{Park2023GenerativeAgents}
Joon~Sung Park, Joseph~C. O'Brien, Carrie~J. Cai, Meredith~Ringel Morris, Percy Liang, and Michael~S. Bernstein.
\newblock Generative agents: Interactive simulacra of human behavior.
\newblock In {\em In the 36th Annual ACM Symposium on User Interface Software and Technology (UIST '23)}, UIST '23, New York, NY, USA, 2023. Association for Computing Machinery.

\bibitem{Liu2023AgentBench}
Xiao Liu, Hao Yu, Hanchen Zhang, Yifan Xu, Xuanyu Lei, Hanyu Lai, Yu~Gu, Yuxian Gu, Hangliang Ding, Kai Men, Kejuan Yang, Shudan Zhang, Xiang Deng, Aohan Zeng, Zhengxiao Du, Chenhui Zhang, Shengqi Shen, Tianjun Zhang, Yu~Su, Huan Sun, Minlie Huang, Yuxiao Dong, and Jie Tang.
\newblock Agentbench: Evaluating llms as agents.
\newblock {\em ArXiv}, abs/2308.03688, 2023.

\bibitem{xie2024travelplanner}
Jian Xie, Kai Zhang, Jiangjie Chen, Tinghui Zhu, Renze Lou, Yuandong Tian, Yanghua Xiao, and Yu~Su.
\newblock Travelplanner: A benchmark for real-world planning with language agents.
\newblock {\em arXiv preprint arXiv:2402.01622}, 2024.

\bibitem{laffont1997game}
Jean-Jacques Laffont.
\newblock Game theory and empirical economics: The case of auction data.
\newblock {\em European Economic Review}, 41(1):1--35, 1997.

\bibitem{tuyls2007evolutionary}
Karl Tuyls and Simon Parsons.
\newblock What evolutionary game theory tells us about multiagent learning.
\newblock {\em Artificial Intelligence}, 171(7):406--416, 2007.

\bibitem{wei2022chain}
Jason Wei, Xuezhi Wang, Dale Schuurmans, Maarten Bosma, brian ichter, Fei Xia, Ed~H. Chi, Quoc~V Le, and Denny Zhou.
\newblock Chain of thought prompting elicits reasoning in large language models.
\newblock In Alice~H. Oh, Alekh Agarwal, Danielle Belgrave, and Kyunghyun Cho, editors, {\em Advances in Neural Information Processing Systems}, 2022.

\bibitem{yao2022react}
Shunyu Yao, Jeffrey Zhao, Dian Yu, Nan Du, Izhak Shafran, Karthik~R Narasimhan, and Yuan Cao.
\newblock React: Synergizing reasoning and acting in language models.
\newblock In {\em Proc. of ICLR}, 2022.

\bibitem{mnih2013playing}
Volodymyr Mnih, Koray Kavukcuoglu, David Silver, Alex Graves, Ioannis Antonoglou, Daan Wierstra, and Martin Riedmiller.
\newblock Playing atari with deep reinforcement learning.
\newblock {\em arXiv preprint arXiv:1312.5602}, 2013.

\bibitem{silver2016mastering}
David Silver, Aja Huang, Chris~J Maddison, Arthur Guez, Laurent Sifre, George Van Den~Driessche, Julian Schrittwieser, Ioannis Antonoglou, Veda Panneershelvam, Marc Lanctot, et~al.
\newblock Mastering the game of go with deep neural networks and tree search.
\newblock {\em nature}, 529(7587):484--489, 2016.

\bibitem{silver2017mastering}
David Silver, Julian Schrittwieser, Karen Simonyan, Ioannis Antonoglou, Aja Huang, Arthur Guez, Thomas Hubert, Lucas Baker, Matthew Lai, Adrian Bolton, et~al.
\newblock Mastering the game of go without human knowledge.
\newblock {\em nature}, 550(7676):354--359, 2017.

\bibitem{yao2023tree}
Shunyu Yao, Dian Yu, Jeffrey Zhao, Izhak Shafran, Thomas~L Griffiths, Yuan Cao, and Karthik Narasimhan.
\newblock Tree of thoughts: Deliberate problem solving with large language models.
\newblock {\em arXiv preprint arXiv:2305.10601}, 2023.

\bibitem{shinn2023reflexion}
Noah Shinn, Federico Cassano, Beck Labash, Ashwin Gopinath, Karthik Narasimhan, and Shunyu Yao.
\newblock Reflexion: Language agents with verbal reinforcement learning.
\newblock {\em arXiv preprint arXiv:2303.11366}, 2023.

\bibitem{schick2023toolformer}
Timo Schick, Jane Dwivedi-Yu, Roberto Dess{\`\i}, Roberta Raileanu, Maria Lomeli, Luke Zettlemoyer, Nicola Cancedda, and Thomas Scialom.
\newblock Toolformer: Language models can teach themselves to use tools.
\newblock {\em arXiv preprint arXiv:2302.04761}, 2023.

\bibitem{hao2023toolkengpt}
Shibo Hao, Tianyang Liu, Zhen Wang, and Zhiting Hu.
\newblock Toolken{GPT}: Augmenting frozen language models with massive tools via tool embeddings.
\newblock In {\em Thirty-seventh Conference on Neural Information Processing Systems}, 2023.

\bibitem{bratman1987intention}
Michael Bratman.
\newblock Intention, plans, and practical reason.
\newblock 1987.

\bibitem{georgeff1999belief}
Michael Georgeff, Barney Pell, Martha Pollack, Milind Tambe, and Michael Wooldridge.
\newblock The belief-desire-intention model of agency.
\newblock In {\em Intelligent Agents V: Agents Theories, Architectures, and Languages: 5th International Workshop, ATAL’98 Paris, France, July 4--7, 1998 Proceedings 5}, pages 1--10. Springer, 1999.

\bibitem{cote2019textworld}
Marc-Alexandre C{\^o}t{\'e}, Akos K{\'a}d{\'a}r, Xingdi Yuan, Ben Kybartas, Tavian Barnes, Emery Fine, James Moore, Matthew Hausknecht, Layla El~Asri, Mahmoud Adada, et~al.
\newblock Textworld: A learning environment for text-based games.
\newblock In {\em Computer Games: 7th Workshop, CGW 2018, Held in Conjunction with the 27th International Conference on Artificial Intelligence, IJCAI 2018, Stockholm, Sweden, July 13, 2018, Revised Selected Papers 7}, pages 41--75. Springer, 2019.

\bibitem{o2023hoodwinked}
Aidan O'Gara.
\newblock Hoodwinked: Deception and cooperation in a text-based game for language models.
\newblock {\em arXiv preprint arXiv:2308.01404}, 2023.

\bibitem{wang2023voyager}
Guanzhi Wang, Yuqi Xie, Yunfan Jiang, Ajay Mandlekar, Chaowei Xiao, Yuke Zhu, Linxi Fan, and Anima Anandkumar.
\newblock Voyager: An open-ended embodied agent with large language models.
\newblock {\em arXiv preprint arXiv: Arxiv-2305.16291}, 2023.

\bibitem{chalamalasetti2023clembench}
Kranti Chalamalasetti, Jana G{\"o}tze, Sherzod Hakimov, Brielen Madureira, Philipp Sadler, and David Schlangen.
\newblock clembench: Using game play to evaluate chat-optimized language models as conversational agents.
\newblock {\em arXiv preprint arXiv:2305.13455}, 2023.

\bibitem{duan2024gtbench}
Jinhao Duan, Renming Zhang, James Diffenderfer, Bhavya Kailkhura, Lichao Sun, Elias Stengel-Eskin, Mohit Bansal, Tianlong Chen, and Kaidi Xu.
\newblock Gtbench: Uncovering the strategic reasoning limitations of llms via game-theoretic evaluations.
\newblock {\em arXiv preprint arXiv:2402.12348}, 2024.

\bibitem{yao2022webshop}
Shunyu Yao, Howard Chen, John Yang, and Karthik Narasimhan.
\newblock Webshop: Towards scalable real-world web interaction with grounded language agents.
\newblock {\em Advances in Neural Information Processing Systems}, 35:20744--20757, 2022.

\bibitem{deng2023mindweb}
Xiang Deng, Yu~Gu, Boyuan Zheng, Shijie Chen, Samuel Stevens, Boshi Wang, Huan Sun, and Yu~Su.
\newblock Mind2web: Towards a generalist agent for the web.
\newblock In {\em Thirty-seventh Conference on Neural Information Processing Systems Datasets and Benchmarks Track}, 2023.

\bibitem{zhou2023webarena}
Shuyan Zhou, Frank~F Xu, Hao Zhu, Xuhui Zhou, Robert Lo, Abishek Sridhar, Xianyi Cheng, Yonatan Bisk, Daniel Fried, Uri Alon, et~al.
\newblock Webarena: A realistic web environment for building autonomous agents.
\newblock {\em arXiv preprint arXiv:2307.13854}, 2023.

\bibitem{yang2023appagent}
Zhao Yang, Jiaxuan Liu, Yucheng Han, Xin Chen, Zebiao Huang, Bin Fu, and Gang Yu.
\newblock Appagent: Multimodal agents as smartphone users.
\newblock {\em arXiv preprint arXiv:2312.13771}, 2023.

\bibitem{chen2023agentverse}
Weize Chen, Yusheng Su, Jingwei Zuo, Cheng Yang, Chenfei Yuan, Chen Qian, Chi-Min Chan, Yujia Qin, Yaxi Lu, Ruobing Xie, et~al.
\newblock Agentverse: Facilitating multi-agent collaboration and exploring emergent behaviors in agents.
\newblock {\em arXiv preprint arXiv:2308.10848}, 2023.

\bibitem{agashe2023evaluating}
Saaket Agashe, Yue Fan, and Xin~Eric Wang.
\newblock Evaluating multi-agent coordination abilities in large language models.
\newblock {\em arXiv preprint arXiv:2310.03903}, 2023.

\bibitem{qian2023communicative}
Chen Qian, Xin Cong, Cheng Yang, Weize Chen, Yusheng Su, Juyuan Xu, Zhiyuan Liu, and Maosong Sun.
\newblock Communicative agents for software development.
\newblock {\em arXiv preprint arXiv:2307.07924}, 2023.

\bibitem{zhao2023competeai}
Qinlin Zhao, Jindong Wang, Yixuan Zhang, Yiqiao Jin, Kaijie Zhu, Hao Chen, and Xing Xie.
\newblock Competeai: Understanding the competition behaviors in large language model-based agents.
\newblock {\em arXiv preprint arXiv:2310.17512}, 2023.

\bibitem{gu2024agent}
Zhouhong Gu, Xiaoxuan Zhu, Haoran Guo, Lin Zhang, Yin Cai, Hao Shen, Jiangjie Chen, Zheyu Ye, Yifei Dai, Yan Gao, et~al.
\newblock Agent group chat: An interactive group chat simulacra for better eliciting collective emergent behavior.
\newblock {\em arXiv preprint arXiv:2403.13433}, 2024.

\bibitem{gemp2024states}
Ian Gemp, Yoram Bachrach, Marc Lanctot, Roma Patel, Vibhavari Dasagi, Luke Marris, Georgios Piliouras, and Karl Tuyls.
\newblock States as strings as strategies: Steering language models with game-theoretic solvers.
\newblock {\em arXiv preprint arXiv:2402.01704}, 2024.

\bibitem{jacob2024modeling}
Athul~Paul Jacob, Abhishek Gupta, and Jacob Andreas.
\newblock Modeling boundedly rational agents with latent inference budgets.
\newblock In {\em The Twelfth International Conference on Learning Representations}, 2024.

\bibitem{huang2024far}
Jen-tse Huang, Eric~John Li, Man~Ho Lam, Tian Liang, Wenxuan Wang, Youliang Yuan, Wenxiang Jiao, Xing Wang, Zhaopeng Tu, and Michael~R Lyu.
\newblock How far are we on the decision-making of llms? evaluating llms' gaming ability in multi-agent environments.
\newblock {\em arXiv preprint arXiv:2403.11807}, 2024.

\bibitem{andreas-2022-language}
Jacob Andreas.
\newblock Language models as agent models.
\newblock In {\em Findings of the Association for Computational Linguistics: EMNLP 2022}, pages 5769--5779, Abu Dhabi, United Arab Emirates, December 2022. Association for Computational Linguistics.

\bibitem{kojima2022large}
Takeshi Kojima, Shixiang~Shane Gu, Machel Reid, Yutaka Matsuo, and Yusuke Iwasawa.
\newblock Large language models are zero-shot reasoners.
\newblock In Alice~H. Oh, Alekh Agarwal, Danielle Belgrave, and Kyunghyun Cho, editors, {\em Advances in Neural Information Processing Systems}, 2022.

\bibitem{richardson2022breakpoint}
Kyle Richardson, Ronen Tamari, Oren Sultan, Reut Tsarfaty, Dafna Shahaf, and Ashish Sabharwal.
\newblock Breakpoint transformers for modeling and tracking intermediate beliefs.
\newblock {\em arXiv preprint arXiv:2211.07950}, 2022.

\bibitem{kagel1986winner}
John~H Kagel and Dan Levin.
\newblock The winner's curse and public information in common value auctions.
\newblock {\em The American economic review}, pages 894--920, 1986.

\bibitem{openai2022chatgpt}
OpenAI.
\newblock Chatgpt, 2022.

\bibitem{OpenAI2023GPT4TR}
OpenAI.
\newblock {GPT-4} technical report.
\newblock {\em ArXiv}, abs/2303.08774, 2023.

\bibitem{team2023gemini}
Gemini Team, Rohan Anil, Sebastian Borgeaud, Yonghui Wu, Jean-Baptiste Alayrac, Jiahui Yu, Radu Soricut, Johan Schalkwyk, Andrew~M Dai, Anja Hauth, et~al.
\newblock Gemini: a family of highly capable multimodal models.
\newblock {\em arXiv preprint arXiv:2312.11805}, 2023.

\bibitem{jiang2023mistral}
Albert~Q Jiang, Alexandre Sablayrolles, Arthur Mensch, Chris Bamford, Devendra~Singh Chaplot, Diego de~las Casas, Florian Bressand, Gianna Lengyel, Guillaume Lample, Lucile Saulnier, et~al.
\newblock Mistral 7b.
\newblock {\em arXiv preprint arXiv:2310.06825}, 2023.

\bibitem{mistralai_2023_mixtral}
{Mistral AI team}.
\newblock Mixtral of experts, December 2023.
\newblock Accessed: 2023-12-15.

\bibitem{NIPS2006_f44ee263}
Ralf Herbrich, Tom Minka, and Thore Graepel.
\newblock Trueskill\texttrademark : A bayesian skill rating system.
\newblock In B.~Sch\"{o}lkopf, J.~Platt, and T.~Hoffman, editors, {\em Advances in Neural Information Processing Systems}, volume~19. MIT Press, 2006.

\bibitem{minka2018trueskill}
Tom Minka, Ryan Cleven, and Yordan Zaykov.
\newblock Trueskill 2: An improved bayesian skill rating system.
\newblock Technical Report MSR-TR-2018-8, Microsoft, March 2018.

\end{thebibliography}

\appendix
\begin{appendix}
\label{sec:appendix}

\section{Examples: Prompt and Bidding History}

\subsection{Bidding History}
\label{appendix:log}

An example of the bidding history is given in List~\ref{lst:bidding_history}, which skips the reasoning process of bidders and only presents the outcomes.

\lstset{
    backgroundcolor=\color[RGB]{245,245,244},
    breaklines=true,
    breakindent=0pt,
    basicstyle=\ttfamily\small,
    emph={Auction,Log,Personal,Report},
    emphstyle={\bfseries\color{brown}}
}\begin{lstlisting}[caption={An example auction log of a bidding war between three bidders over three items.},label=lst:bidding_history]
## Auction Log

### 1. Gadget B, starting at $1000.

#### 1st bid:

* Bidder 1: $1200
* Bidder 2: $1000
* None bid

#### 2nd bid:

* Bidder 1: $1300
* Bidder 2: Withdrew

#### 3rd bid:

* Bidder 3: $1400

#### 4th bid:

* Bidder 1: $1500

#### 5th bid:

* Bidder 3: Withdrew

#### Hammer price (true value):

* Bidder 1: $1500 ($2000)

### 2. Thingamajig C, starting at $1000.

#### 1st bid:

* Bidder 1: $1200
* Bidder 2: $1100
* Bidder 3: $2000

#### 2nd bid:

* Bidder 1: Withdrew
* Bidder 2: Withdrew

#### Hammer price (true value):

* Bidder 3: $2000 ($2000)

### 3. Widget A, starting at $1000.

#### 1st bid:

* Bidder 1: $1100
* Bidder 2: $1200
* Bidder 3: $1100

#### 2nd bid:

* Bidder 1: Withdrew
* Bidder 3: Withdrew

#### Hammer price (true value):

* Bidder 2: $1200 ($2000)


## Personal Report

* Bidder 1, starting with $20000, has won 1 items in this auction, with a total profit of $500.:
  * Won Gadget B at $1500 over $1000, with a true value of $2000.

* Bidder 2, starting with $20000, has won 1 items in this auction, with a total profit of $800.:
  * Won Widget A at $1200 over $1000, with a true value of $2000.

* Bidder 3, starting with $0, has won 1 items in this auction, with a total profit of $0.:
  * Won Thingamajig C at $2000 over $1000, with a true value of $2000.
  
\end{lstlisting}

\subsection{Instruction Prompts}
\label{appendix:prompt}

A complete bidder agent has these functions: planning, bidding, belief update, replanning ($\mathsection$~\ref{sec:agent}).
Here, we report the arguments of each function as follows:
\begin{itemize}
    \item Planning: System Message, Initial Belief, Planning Instruction;
    \item Bidding: System Message, (Re-)Plan Instruction, (Updated) Plan, Bidding History, Bid Instruction;
    \item Belief Update: System Message, Bidding History, Belief Update Instruction;
    \item Replanning: System Message, Updated Belief, Replanning Instruction;
\end{itemize}
The System Message and Planning, Bidding, Belief Update, Replanning instructions are shown in Listing~\ref{lst:sample}.
We keep the instructions as general as possible, providing only the necessary rules of the auction and examples of output format for parsing.
We try to not provide any examples of concrete auction strategies for in-context learning to avoid any unintended biases.
Due to budget limits, we do not have the resources to rigorously evaluate more forms of instruction designs like in the experiments, other than some prompt engineering endeavors during the development period.

\lstset{
    backgroundcolor=\color[RGB]{245,245,244},
    breaklines=true,
    breakindent=0pt,
    basicstyle=\ttfamily\small,
    emph={System,Message,Planning,Instruction,Bidding,Belief,Update,Replanning},
    emphstyle={\bfseries\color{brown}}
}\begin{lstlisting}[caption={The System Message and Planning, Bidding, Belief Update, and Replanning Instructions.},label=lst:sample]
System Message:

You are {name}, who is attending an ascending-bid auction as a bidder. This auction will have some other bidders to compete with you in bidding wars. The price is gradually raised, bidders drop out until finally only one bidder remains, and that bidder wins the item at this final price. Remember: Your primary objective is to secure the highest profit at the end of this auction, compared to all other bidders.

Here are some must-know rules for this auction:

1. Item Values: The true value of an item means its resale value in the broader market, which you don't know. You will have a personal estimation of the item value. However, note that your estimated value could deviate from the true value, due to your potential overestimation or underestimation of this item.
2. Winning Bid: The highest bid wins the item. Your profit from winning an item is determined by the difference between the item's true value and your winning bid. You should try to win an item at a bid as minimal as possible to save your budget.

Planning Instruction:

As {bidder_name}, you have a total budget of ${budget}. This auction has a total of {item_num} items to be sequentially presented, they are:
{items_info}

------------------------------

Please plan for your bidding strategy for the auction based on the information{learning_statement}. A well-thought-out plan positions you advantageously against competitors, allowing you to allocate resources effectively. With a clear strategy, you can make decisions rapidly and confidently, especially under the pressure of the auction environment. Remember: Your primary objective is to secure the highest profit at the end of this auction, compared to all other bidders. 

After articulate your thinking, in you plan, assign a priority level to each item. Present the priorities for all items in a JSON format, each item should be represented as a key-value pair, where the key is the item name and the value is its priority on the scale from 1-3. An example output is: {{"Fixture Y": 3, "Module B": 2, "Product G": 2}}. The descriptions of the priority scale of items are as follows.
    * 1 - This item is the least important. Consider giving it up if necessary to save money for the rest of the auction.
    * 2 - This item holds value but isn't a top priority for the bidder. Could bid on it if you have enough budget.
    * 3 - This item is of utmost importance and is a top priority for the bidder in the rest of the auction.

Bidding Instruction:

Now, the auctioneer says: "Attention, bidders! {num_remaining_items} item(s) left, they are: {item_info}. Now, please bid on {cur_item}. The starting price for bidding for {cur_item} is ${cur_item_price}. Anyone interested in this item?" / "Thank you! This is the {bid_round} round of bidding for this item: {bidding_history}. Now we have ${highest_bid} from {highest_bidder} for {cur_item}. The minimum increase over this highest bid is ${min_increse}. Do I have any advance on ${highest_bid}?"

------------------------------

As {bidder_name}, you have to decide whether to bid on this item or withdraw and explain why, according to your plan{learning_statement}. Remember, Your primary objective is to secure the highest profit at the end of this auction, compared to all other bidders.

Here are some common practices of bidding:
1. Showing your interest by bidding with or slightly above the starting price of this item, then gradually increase your bid.
2. Think step by step of the pros and cons and the consequences of your action (e.g., remaining budget in future bidding) in order to achieve your primary objective.

Give your reasons first, then make your final decision clearly. You should either withdraw (saying "I'm out!") or make a higher bid for this item (saying "I bid $xxx!").

Belief Update Instruction:

Here is the history of the bidding war of {cur_item}:
"{bidding_history}"

The auctioneer concludes: "{hammer_msg}" 

------------------------------

Congratulations! You have won {item} at {bid_price} / You have lost {item}.
As {bidder_name}, you have to update the status of the auction based on this round of bidding. Here's your previous status:
```
{prev_status}
```

Summarize the notable behaviors of all bidders in this round of bidding for future reference. Then, update the status JSON regarding the following information:
- 'remaining_budget': The remaining budget of you, expressed as a numerical value.
- 'total_profits': The total profits achieved so far for each bidder, where a numerical value following a bidder's name. No equation is needed, just the numerical value.
- 'winning_bids': The winning bids for every item won by each bidder, listed as key-value pairs, for example, {{"bidder_name": {{"item_name_1": winning_bid}}, {{"item_name_2": winning_bid}}, ...}}. If a bidder hasn't won any item, then the value for this bidder should be an empty dictionary {{}}.
- Only include the bidders mentioned in the given text. If a bidder is not mentioned (e.g. Bidder 4 in the following example), then do not include it in the JSON object.

After summarizing the bidding history, you must output the current status in a parsible JSON format. An example output looks like:
```
{{"remaining_budget": 8000, "total_profits": {{"Bidder 1": 1300, "Bidder 2": 1800, "Bidder 3": 0}}, "winning_bids": {{"Bidder 1": {{"Item 2": 1200, "Item 3": 1000}}, "Bidder 2": {{"Item 1": 2000}}, "Bidder 3": {{}}}}}}
```

Replanning Instruction:

The current status of you and other bidders is as follows:
```
{status_quo}
```

Here are the remaining items in the rest of the auction:
"{remaining_items_info}"

As {bidder_name}, considering the current status{learning_statement}, review your strategies. Adjust your plans based on the outcomes and new information to achieve your primary objective. This iterative process ensures that your approach remains relevant and effective. Please do the following:
1. Always remember: Your primary objective is to secure the highest profit at the end of this auction, compared to all other bidders.
2. Determine and explain if there's a need to update the priority list of remaining items based on the current status. 
3. Present the updated priorities in a JSON format, each item should be represented as a key-value pair, where the key is the item name and the value is its priority on the scale from 1-3. An example output is: {{"Fixture Y": 3, "Module B": 2, "Product G": 2}}. The descriptions of the priority scale of items are as follows.
    * 1 - This item is the least important. Consider giving it up if necessary to save money for the rest of the auction.
    * 2 - This item holds value but isn't a top priority for the bidder. Could bid on it if you have enough budget.
    * 3 - This item is of utmost importance and is a top priority for the bidder in the rest of the auction.
\end{lstlisting}

\subsection{Information of Artificial Items in the Auction}
\label{appendix:items}

\begin{table*}[t]
\centering
\small
\begin{tabular}{cccc}
\toprule
\textbf{Name} & \textbf{Price} & \textbf{Description}                          & \textbf{True Value} \\ 
\midrule
Widget A      & \$1000           & A widget for all your needs                   & \$2000                \\ \hline
Gadget B      & \$3000           & A gadget with all the latest features         & \$6000                \\ \hline
Thingamajig C & \$4000           & A little thing that is sure to impress        & \$8000                \\ \hline
Doodad D      & \$2000           & A durable doodad that will last for years     & \$4000                \\ \hline
Equipment E   & \$5000           & A piece of equipment for any tough job        & \$10000               \\ \hline
Gizmo F       & \$3000           & A gizmo that will surprise and delight        & \$6000                \\ \hline
Implement G   & \$2000           & A implement for everyday tasks                & \$4000                \\ \hline
Apparatus H   & \$4000           & An apparatus for specialized operations       & \$8000                \\ \hline
Contraption I & \$1000           & A contraption that sparks creativity          & \$2000                \\ \hline
Mechanism J   & \$5000           & A mechanism for repetitive tasks              & \$10000               \\ \bottomrule
\end{tabular}
\caption{Overview of artificial items in the experiments.}
\label{tab:items}
\end{table*}

In Table~\ref{tab:items}, we list the information of every item.
We use artificial item names and descriptions to avoid any unintentional information leakage that may influence the results.
In the auction, the order of items might be changed, \eg, items might be randomly shuffled, or sorted in an ascending or descending order according to their starting prices and true values.

\section{Additional Results}

\begin{table*}[t]
\small
\centering
\begin{tabular}{ccccccc}
\toprule
\multirow{2}{*}{\textbf{Bidder}} & \multicolumn{2}{c}{\textbf{Random}} & \multicolumn{2}{c}{\textbf{Ascending}} & \multicolumn{2}{c}{\textbf{Descending}} \\
& \textbf{Profit (\$)} & \textbf{\# Items} & \textbf{Profit (\$)}& \textbf{\# Items} & \textbf{Profit (\$)}& \textbf{\# Items} \\
\midrule
GPT-4-Turbo & \textbf{8712.5} & \textbf{5.0} & 7525.0 & \textbf{5.0} & 8287.5 & 4.3 \\
\midrule
GPT-3.5-Turbo & 3325.0 & 1.0 & 2875.0 & 1.3 & 950.0 & 0.5 \\
\midrule
Human & 7237.5 & 3.8 & \textbf{10862.5} & 3.8 & \textbf{10025.0} & \textbf{5.0} \\
\midrule
\end{tabular}
\caption{Comparison of averaged profits and the number of won items across different bidders.}
\label{tab:human}
\end{table*}

\subsection{Human Performance}
\label{appendix:human}
We incorporated human evaluation by recruiting 4 human players to participate in the auction in the experiment with item orders, who are graduate students and paid above the local minimum wage. 
As with the same setup described in $\mathsection$~\ref{sec:benchmarking}, we set a budget of \$20,000 and have the auctioneer present 10 items. 
The models that participate in the bidding are a GPT-4-Turbo, a GPT-3.5-Turbo, and a human bidder. The following is the average profit and the number of items won by each agent in this 3-player bidding (note that the human players are not the same individual, so TrueSkill is not suitable here; we use the average profit for comparison).

The results in Table~\ref{tab:human} show that, in the Ascending and Descending scenarios, humans often achieved better profits due to their more accurate estimation of item values in the long run. 
Humans have a clearer strategy of saving money for valuable items in the Ascending scenario and bidding more aggressively for high-value items in the Descending scenario (or you will miss the chance). 
GPT-4-Turbo actively participates in bidding but is not as aggressive as the human players in the Descending scenario, GPT-3.5-Turbo, on the other hand, shows frequent withdrawal from bidding (even when the budget is ample). 

However, in the Random scenario, GPT-4-Turbo's average profit even exceeded that of human players.
This indicates that humans still have a harder time making reasonable strategies when faced with unpredictable item values. While humans excel in structured scenarios where a clear pattern can be discerned, their performance declines in random settings where adaptability and quick decision-making are crucial. In contrast, GPT-4-Turbo’s consistent performance across different scenarios suggests a balanced approach to bidding, which pays off in unpredictable environments.

Additionally, we notice several common errors in the models after case study, which is presented in List~\ref{list:case_study}:
1) GPT-4-Turbo gives reasonable bids but lacks advanced strategies to limit opponents with minimal increases and drain their budgets.
2) GPT-3.5-Turbo frequently gives up bidding, failing to utilize its budget for greater benefits.

\

\lstset{
    backgroundcolor=\color[RGB]{245,245,244},
    breaklines=true,
    breakindent=0pt,
    basicstyle=\ttfamily\small,
    emph={GPT-3.5,GPT-4,Case 1, Case 2},
    emphstyle={\bfseries\color{brown}}
}\begin{lstlisting}[caption={Case study of GPT-3.5 and GPT-4 bidder in the arena.},label=list:case_study]
GPT-4:

Case 1:
Considering that Contraption I has a starting price of $1000 and my estimated value for this item is $2200, there is a potential profit margin of $1200 if my estimation is accurate. Since we have yet to win any items and our budget remains intact at $20,000, this could be a good opportunity to begin participating in the auction.
However, we must be cautious not to get caught up in a bidding war that would significantly reduce this potential profit margin. It's important to keep in mind that there are other items down the line that may offer a better profit opportunity, and we want to ensure we have enough budget to compete for those items as well.
Here are the pros and cons of bidding on Contraption I:
Pros:
Potential profit margin of $1200 based on estimation.
Bidding could signal to other bidders our readiness to participate, possibly deterring aggressive bids on later items.
Cons:
Risk of overbidding if competition is fierce, which could reduce our overall profit.
Winning this item at a high price could deplete our budget for future items with potentially higher profit margins.
Considering these factors, a cautious initial bid that is close to the starting price could be a strategic move. It shows interest without committing too much capital upfront. If the price escalates quickly beyond a comfortable margin, we can always decide to withdraw.
Final Decision:
"I bid $1050!"

Case 2:
Bidder 3 has placed a bid of $2300 on Implement G, which is still well below my estimated value of $4400. Given the potential profit margin of $2100 if I win at this bid, it is still advantageous for me to continue bidding.
The minimum increase is $200, so a bid of $2500 would be appropriate to show my continued interest without pushing the price too high too quickly. This would still leave me with a potential profit margin of $1900, assuming my estimate is accurate.
It's important to remember that other bidders may also recognize the value of Implement G and could be willing to bid higher. I need to be vigilant and ready to step back if the price escalates to a point where the potential profit no longer justifies the risk.
Considering these points, I will place the next bid:
"I bid $2500!"

GPT-3.5:

Case 1:
I will not bid on Implement G. This decision aligns with my strategy to secure the highest profit at the end of the auction. The starting price for Implement G is $2000, and based on my estimated value of $4400, I don't believe it aligns with my priority to prioritize high-value items. Additionally, it's important to conserve my budget for items of higher priority in order to achieve my primary objective.
Therefore, I'm out!

Case 2:
I'm out!
Based on my analysis and the current status of the auction, I have decided to withdraw from bidding on Equipment E. Although it is a high-value item with an estimated value of $11000, my remaining budget of $15000 is not sufficient to make a competitive bid for this item. Considering the potential bidding wars and the need to strategically manage my budget for future bidding, it is not in my best interest to pursue this item at the current starting price of $5000.
Therefore, in line with my primary objective of securing the highest profit at the end of this auction, I have made the decision to withdraw from bidding on Equipment E.

\end{lstlisting}

\subsection{Visualizing Long-term Strategies: A Budget of \$20,000}
\label{appendix:heatmap}

\begin{figure*}[t]
    \centering
    \small
    \includegraphics[width=\linewidth]{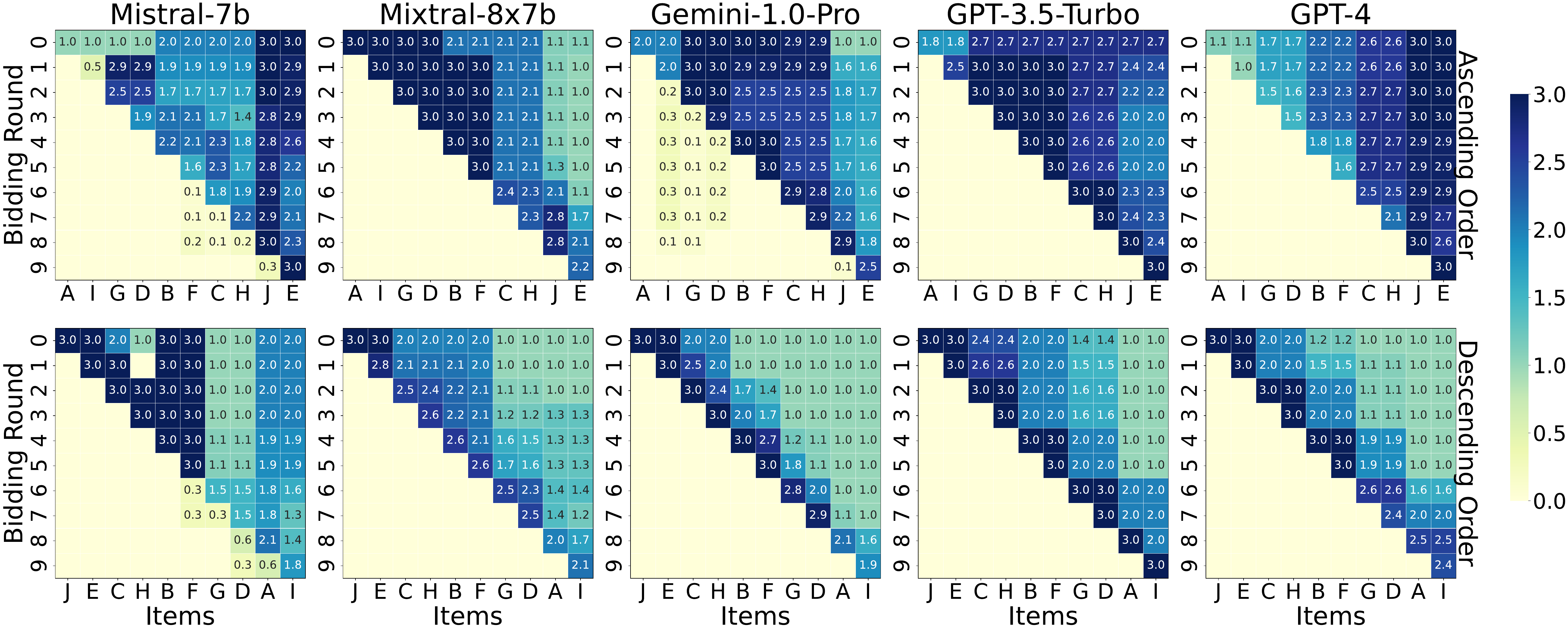}
    \caption{The average heatmap of priority changes after each replanning step when the budget is \$40,000. Note that the priority score ranges from 1 to 3.}
    \label{fig:heatmap_40k}
\end{figure*}

As shown in Figure~\ref{fig:heatmap_40k}, we visualize the priority score changes in the plan for all models when the budget is \$40,000. 
The results echo those of Figure~\ref{fig:heatmap} (budget: \$20,000), most LLMs demonstrate reasonable strategic planning capabilities and can conserve budgets for the future.
When items are presented in an ascending order, different models present different preferences.
Some prefer medium-value items (Gemini-1.0-Pro), and some prefer high-value ones (GPT-4, Mistral-7b).
Curiously, both Gemini-1.0-Pro and Mistral-7b exhibit similar planning errors and show outliers in priority scores.
They have this error by allocating priority to past items, which somehow shows their inferior state tracking and instruction following abilities.
While models typically focus their bidding on the current item in a descending order due to subsequent items being less costly and offering lower profit margins, they tend to raise priority scores for less expensive items. 
This is because, when the auction moves to low-value items, not many options are left for bidders to make profit.
Thus, agents would prioritize winning these as every dollar is significant now.

\subsection{Execution Analysis: Visualizing Bidding Behaviors}
\label{appendix:bubble}

\begin{figure*}[t]
    \centering
    \includegraphics[width=\linewidth]{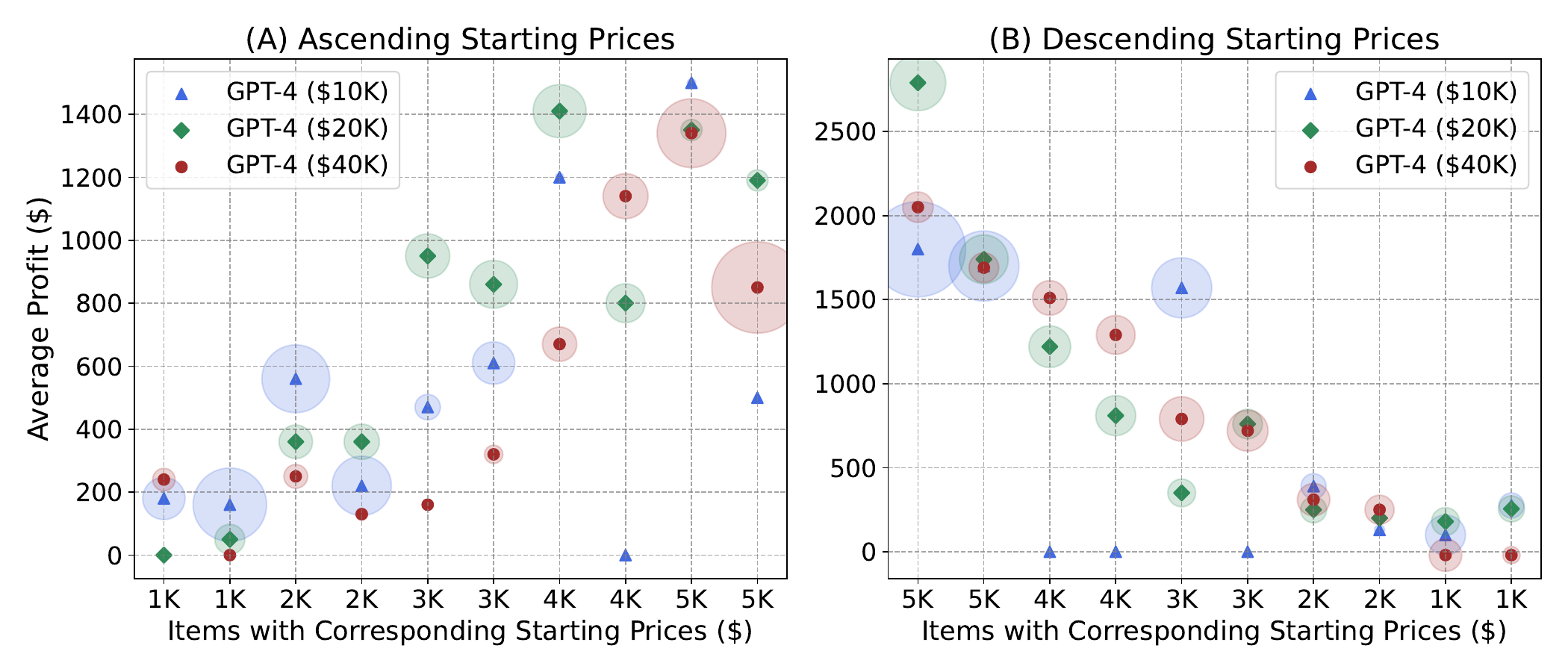}
    \caption{The bidding behaviors of GPT-4 influenced by the starting prices and orders of items in standard competition for three budget settings (\$10,000, \$20,000, \$40,000). 
    On the left (A), items are presented with ascending starting prices on the x-axis from \$1,000 to \$5,000, while the order is descending on the right (B). 
    The y-axis shows the average profit from winning the corresponding item. 
    The size of each circle represents the average number of bids placed on the item. }
    \label{fig:bubble}
\end{figure*}

\begin{figure*}[t]
    \centering
    \includegraphics[width=\linewidth]{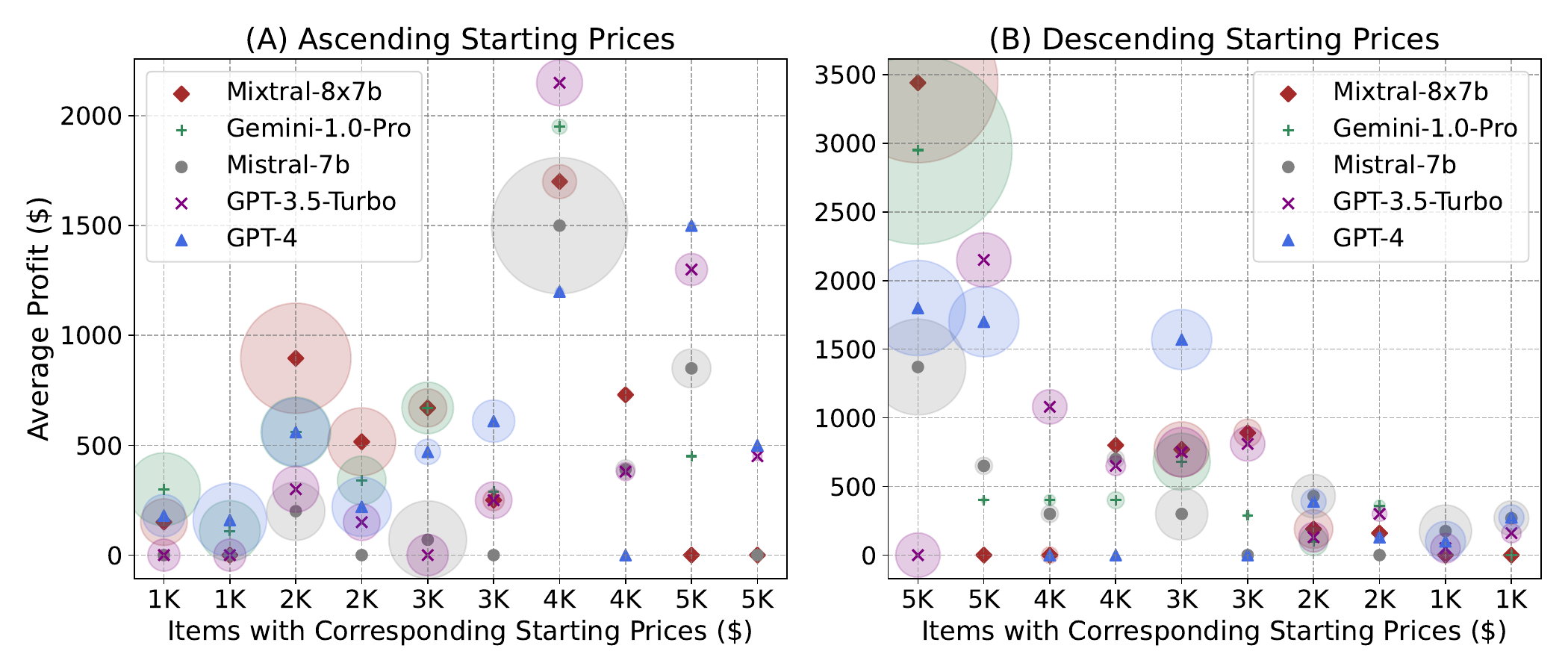}
    \caption{The bidding behaviors under a budget of \$10,000 influenced by the starting prices and orders of items in the standard competition.}
    \label{fig:bubble_10k}
\end{figure*}

\begin{figure*}[t]
    \centering
    \includegraphics[width=\linewidth]{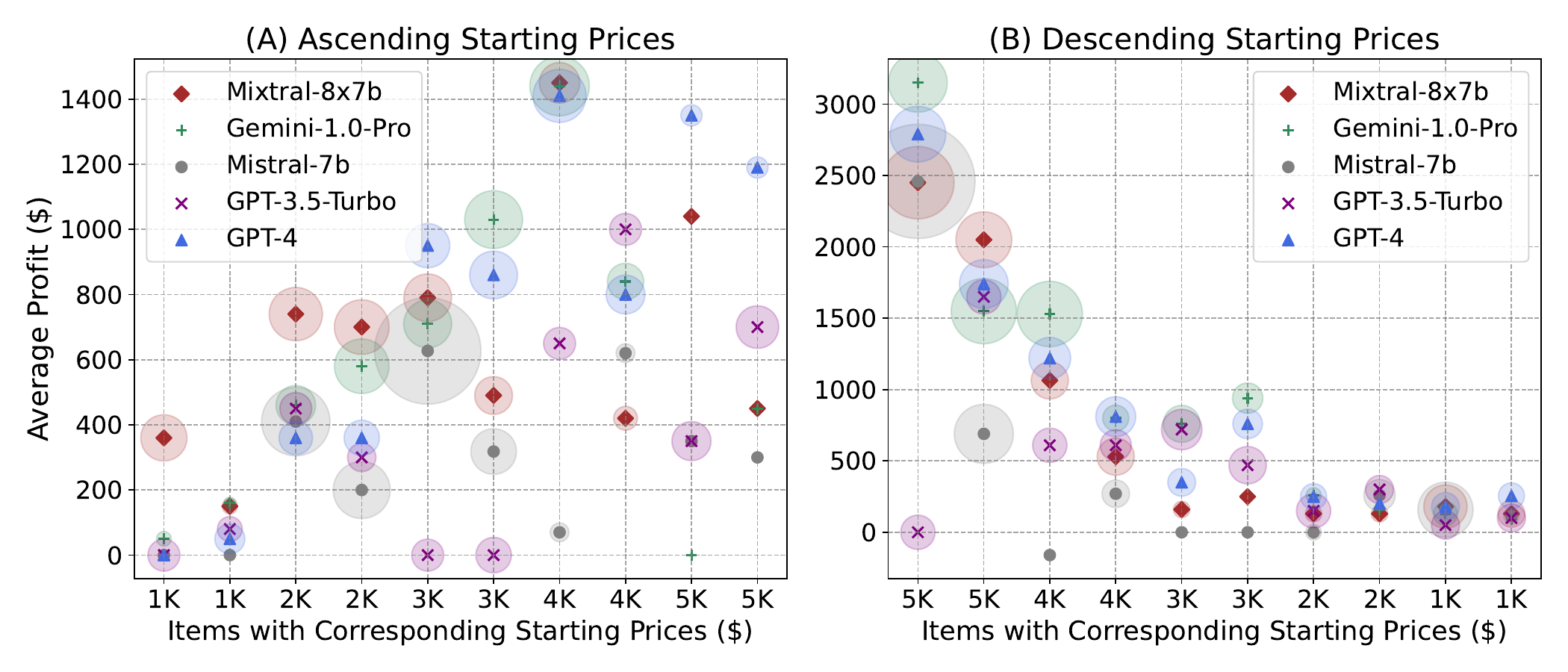}
    \caption{The bidding behaviors under a budget of \$20,000 influenced by the starting prices and orders of items in the standard competition.}
    \label{fig:bubble_20k}
\end{figure*}

\begin{figure*}[t]
    \centering
    \includegraphics[width=\linewidth]{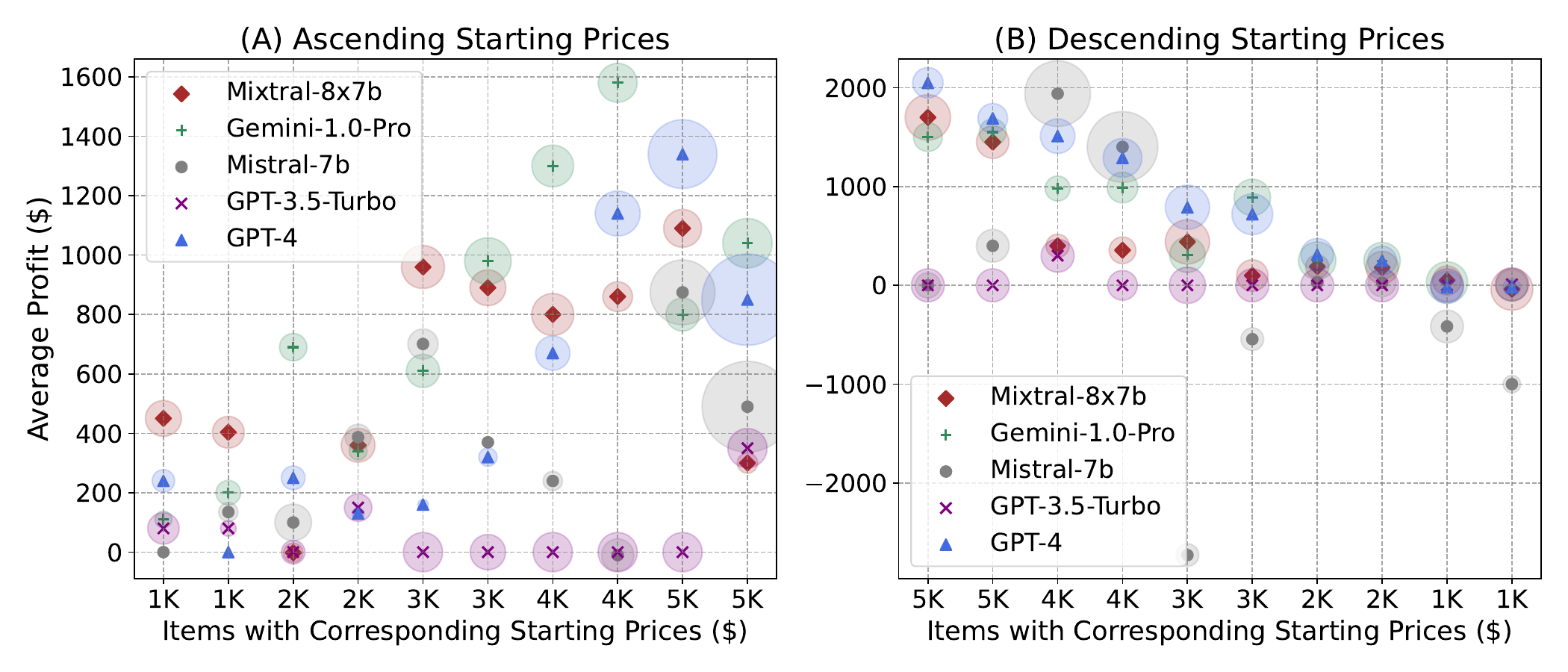}
    \caption{The bidding behaviors under a budget of \$40,000 influenced by the starting prices and orders of items in the standard competition.}
    \label{fig:bubble_40k}
\end{figure*}

We report the results of GPT-4 in Figure~\ref{fig:bubble} where the budget is set as insufficient (\$10,000), sufficient (\$20,000) and abundant (\$40,000), and leave the rest models to Appendix~\ref{appendix:bubble}.
By comparing cases where item orders are different (ascending or descending starting prices), we can observe the bidding patterns of LLMs (bidding frequency) on specific items and the profits they generate.
In Figure~\ref{fig:bubble}, GPT-4's bidding behavior with varying budgets reveals different strategies. 
With a \$10,000 budget, bids are cautious across items. 
As the budget increases to \$20,000 and \$40,000, bids and profits rise, especially for items in the mid-price and high-price range in ascending order (A). 
In descending order (B), the lower budget prompts aggressive bidding on high-priced items, while one with the highest budget shows a more even distribution of bids.
This suggests GPT-4 adapts its strategy based on budget size and item order, becoming more competitive for higher-value items when budgets allow.
However, other models do not share the same behaviors.
For example, GPT-3.5-Turbo consistently bids at a unanimous frequency across all items in all budget settings, as evidenced by Appendix~\ref{appendix:bubble}.

To supplement the GPT-4 results, we further visualize the bidding behaviors of five LLMs (Mixtral-7b, Mixtral-8x7b, Gemini-1.0-Pro, GPT-3.5-Turbo and GPT-4) when setting the budget as \$10,000 (insufficient), \$20,000 (sufficient), and \$40,000 (abundant), shown in Figure~\ref{fig:bubble_10k},~\ref{fig:bubble_20k},~\ref{fig:bubble_40k}, respectively.

When the budget is insufficient (\$10,000), agents cannot win every item, and more profitable ones also invite more intensive bidding wars.
In Figure~\ref{fig:bubble_10k}, it is noticeable that most agents go hard for first few items, especially when the items are presented in a descending order.
Some exceptions exist, \eg, Mistral-7b fights hard for items with a starting price of \$4,000.
GPT-3.5-Turbo also profits much from \$4,000 items, but it is the result of its greedy strategy, as few bidders have budget left at a later stage to compete with it.
Same reasons go to the winners for items starting with \$5,000, where most wins come with few bids.

However, things are significantly different when the budget is abundant (\$40,000) in Figure~\ref{fig:bubble_40k}.
In the ascending price scenario, agents spend more budget on the more profitable items, which leads to a more intensive competition where more bids are given and the profit for winning is less.
This results in the observation that GPT-3.5-Turbo wins less items, shown by its 0 profit for most items, ascending or descending alike.
In the ascending price case (Figure~\ref{fig:bubble_40k}(A)), most models choose to skip the first two low-value items, while GPT-4 often secures high-value items with few bids at a later stage, indicating its success in patience and long-term strategies.
In the descending cases (Figure~\ref{fig:bubble_40k}(B)), the competition for the first few items is quite intensive, as shown by large circle sizes.
Mistral-7b sometimes overbids for an item, suffering from the winner's curse and losing money for it.

Among these models, GPT-3.5-Turbo consistently bids at a uniform frequency across all items (shown by equal sizes of all bid freqency circles), akin to a rule-based bidder. 
This behavior corroborates the observation in Figure~\ref{fig:heatmap} that GPT-3.5-Turbo lacks a strategic bidding plan, instead opting to place several bids before withdrawing. 
For instance, with an abundant budget, the bidding approach of GPT-3.5-Turbo is less effective, as competitors possess sufficient funds to remain competitive.

\end{appendix}

\end{document}